\documentclass{article} % For LaTeX2e
\usepackage{iclr2021_conference_arxiv,times}

\usepackage{amsmath,amsfonts,bm}

\def\eqref#1{equation~\ref{#1}}
\def\1{\bm{1}}

\DeclareMathAlphabet{\mathsfit}{\encodingdefault}{\sfdefault}{m}{sl}
\SetMathAlphabet{\mathsfit}{bold}{\encodingdefault}{\sfdefault}{bx}{n}

\DeclareMathOperator*{\argmin}{arg\,min}

\usepackage{hyperref}

\usepackage{graphicx}
\usepackage{url}
\usepackage{amsmath, amssymb}
\usepackage{array, booktabs}
\usepackage{comment}
\usepackage{algorithm}
\usepackage{algorithmic}

\title{Generative Language-Grounded Policy in \\
Vision-and-Language Navigation with Bayes' Rule}

\iclrfinalcopy

\author{Shuhei Kurita$^1$ \thanks{ This work was done when the first author visited New York University.} \\
\texttt{shuhei.kurita@riken.jp~~~~~~~~~kyunghyun.cho@nyu.edu} \\
\And
\hspace{-20.05em}Kyunghyun Cho$^{2,3,4}$ \\
\texttt{} \\
\And
\vspace{-1em}\\
$^1$RIKEN AIP\\
$^2$Center for Data Science, New York University \\
$^3$Department of Computer Science, Courant Institute, New York University \\
$^4$CIFAR Associate Fellow \\
}

\begin{document}

\maketitle

% \begin{abstract}
% The abstract paragraph should be indented 1/2~inch (3~picas) on both left and
% right-hand margins. Use 10~point type, with a vertical spacing of 11~points.
% The word \textsc{Abstract} must be centered, in small caps, and in point size 12. Two
% line spaces precede the abstract. The abstract must be limited to one
% paragraph.
% \end{abstract}

\begin{abstract}
Vision-and-language navigation (VLN) is a task in which an agent is embodied in a realistic 3D environment and follows an instruction to reach the goal node. While most of the previous studies have built and investigated a discriminative approach, we notice that there are in fact two possible approaches to building such a VLN agent: discriminative \textit{and} generative. In this paper, we design and investigate a generative language-grounded policy which uses a language model to compute the distribution over all possible instructions i.e. all possible sequences of vocabulary tokens given action and the transition history. In experiments, we show that the proposed generative approach outperforms the discriminative approach in the Room-2-Room (R2R) and Room-4-Room (R4R) datasets, especially in the unseen environments. We further show that the combination of the generative and discriminative policies achieves close to the state-of-the art results in the R2R dataset, demonstrating that the generative and discriminative policies capture the different aspects of VLN.

\end{abstract}

\section{Introduction}

Vision-and-language navigation \citep{mattersim} is a task in which a computational model follows an instruction and performs a sequence of actions to reach the final objective. An agent is embodied in a realistic 3D environment, such as that from the Matterport 3D Simulator~\cite{Matterport3D} and asked to follow an instruction. The agent observes the surrounding environment and moves around. This embodied agent receives a textual instruction to follow before execution. The success of this task is measured by how accurately and quickly the agent could reach the destination specified in the instruction. VLN is a sequential decision making problem: the embodied agent makes a decision each step considering the current observation, transition history and the initial instruction.

Previous studies address this problem of VLN by building a language grounded policy which computes a distribution over all possible actions given the current state and the language instruction.
In this paper, we notice there are two ways to formulate the relationship between the action and instruction.
%First approach
First, the action is assumed to be generated from the instruction, similarly to most of the existing approaches~\citep{mattersim,vln-sm,vln-rcm,vln-areyoulooking,trl-vln}. This is often called a follower model~\citep{speaker-follower}.
We call it a discriminative approach analogous to logistic regression in binary classification.

On the other hand, the action may be assumed to generate the instruction.
In this case, we build a neural network to compute the distribution over all possible instructions given an action and the transition history.
With this neural network, we use Bayes' rule to build a language-grounded policy.
We call this generative approach,
similarly to na\"ive Bayes in binary classification.
%It is although similar

Despite its similarity
to the speaker model of \cite{speaker-follower}, there is a stark difference 
%different from speaker model because
that the speaker model takes as input the entire sequence of actions and then the predicts the entire instruction, which is not the case in ours.
Instead, the generative language-grounded policy only considers what is available at each time step and chooses one of the potential actions to generate the instruction.
We then apply Bayes' rule to obtain the posterior distribution over actions given the instruction.

Given these discriminative and generative parameterizations of the language-grounded policy, we hypothesize that the generative parameterization works better than discriminative parameterization does, because the former benefits from richer learning signal arising from generating the entire instruction rather than predicting a single action.

\begin{figure}[t]
	\hspace{0em}
    \begin{center}
	\includegraphics[scale=0.75,clip]{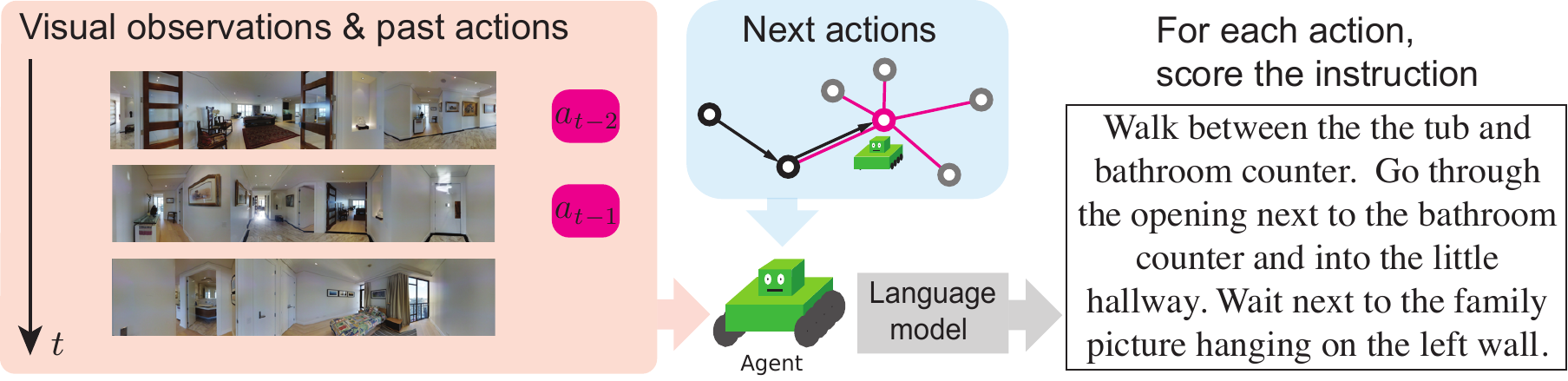}
	\vspace{0em}
      \caption{
			 The generative language-grounded policy for vision-and-language navigation.
			 }
      \label{fig:initialfig}
    \vspace{-1.5em}
    \end{center}
\end{figure}

We empirically show that indeed the proposed generative approach outperforms the discriminative approach in both the R2R and R4R datasets, especially in the \textit{unseen environments}.
Figure~\ref{fig:initialfig} illustrates the proposed generative approach on VLN.
%\footnote{Note that the representations of the actions in the R2R navigation are computed from the directions to move and current visual observation. Our model doesn't exploit any visual observations where the agent hasn't visited, e.g. gray dots in the figure.}
%\footnote{The code is available at \url{ANONYMIZED URL}}
Furthermore, we show that the combination of the generative and discriminative policies results in near state-of-the art results in R2R and R4R, demonstrating that they capture two different aspects of VLN.
%we could say something along the line of:
We demonstrate that the proposed generative policy is more interpretable than the conventional discriminative policy, by introducing a token-level prediction entropy as a way to measure the influence of each token in the instruction on the policy's decision.

\section{Discriminative and Generative Parameterizations of Language-Grounded Policy}

%Humans make decisions based on language, which enables our higher-order planning ability that consists of numerous small and local decision making.
%how local decisions effect the entire planning.

%In this section, we propose a novel method that allows a sequential decisions
%by generating an language instruction for each action.

Vision-and-language navigation (VLN) is a sequential decision making task, where an agent performs a series of actions based on the initially-given instruction, visual features, and past actions.
%At time $t$ the agent observes visual features $s_t$ and takes an action $a_t$.
Given the instruction $X$, past and current observations $s_{:t}$ and past actions $a_{:t-1}$, the agent computes the distribution $p(a_t|X,s_{:t},a_{:t-1})$ at time $t$.
% over the $t$-th action as
For brevity, we write the current state that consists of the current and past scene observations, and past actions as $h_t=\{s_{:t},a_{:t-1}\}$, and the next action prediction as $p(a_t|X,h_t)$.
The instruction $X$ is a sequence of tokens $X=(w_0, w_1,...,w_k,...)$.
%Figure~\ref{fig:modelparams} illustrates the relationship between these notations.
The relationship between these notations are also presented in Appendix~\ref{sec:notations}.

%Especially, for a completed trajectory, we use a notation of $H={s_{*},a_{*}}$ for its past full observation and transition actions.

In VLN, the goal is to model $p(a_t|h_t,X)$ so as to maximize the success rate of reaching the goal while faithfully following the instruction $X$. In doing so, there are two approaches: \textit{generative} and \textit{discriminative}, analogous to solving 
% binary 
classification with either logistic regression or naive Bayes.

In the \textit{discriminative} approach, we build a neural network to directly estimate $p(a_t|h_t,X)$.
This neural network takes as input the current state $h_t$ and the language instruction $X$ and outputs a distribution over the action set. Learning corresponds to
\begin{equation}
\max_\theta \sum_{n=1}^{N} \sum_{t=1}^{T_n} \log p(a_{t}^n|h_t^n,X^n),
\end{equation}
%\label{eq:desc}
where $N$ is the number of training trajectories.

In the \textit{generative} approach, on the other hand, we first rewrite the action distribution as
\begin{equation}
p(a_{t}|h_t,X) = \frac{p(X|a_{t},h_t)p'(a_t|h_t)}{\sum_{a'_t\in  \mathcal{A}}{p(X|a'_t,h_t)p'(a'_t|h_t})} = \frac{p(X|a_{t},h_t)}{\sum_{a'_t \in \mathcal{A}}{p(X|a'_t,h_t)}},
\label{eq:gen}
\end{equation}
assuming $p'(a_t|h_t)=1/|\mathcal{A}|$, where $\mathcal{A}$ is the action set. This assumption implies that the action is independent of the state without the language instruction, which is a reasonable assumption as the goal is specified using the instruction $X$. 
% Under this assumption,
% \begin{equation}
% p(a_{t}|h_t,X) = \frac{p(X|a_{t},h_t)}{\sum_{a'_t \in \mathcal{A}}{p(X|a'_t,h_t)}}.
% \label{eq:gen2}
% \end{equation}
%We then use a neural network to model $p(X|a_t,h_t)$.
$p(X|a_t,h_t)=\Pi_k p(w_k|a_t,h_t,w_{:k-1})$ is a language model conditioned on an action $a_t$ and the current hidden state $h_t$, and outputs the distribution over all possible sequences of vocabulary tokens. 

Learning is then equivalent to solving
\begin{eqnarray}
\max_\theta \sum_{n=1}^{N} \sum_{t=1}^{T_n} \Big( \log p(X^n|a_{t}^n,h_t^n) - \log \sum_{a_t^{\prime n} \in \mathcal{A} } p(X^n|a_t^{\prime n},h_t^n) \Big)
\label{eq:overall}.
\end{eqnarray}
$\log p(X^n|a_t^{n},h_t^n)$ is the language model loss conditioned on the reference action $a_t^n$, while the second term $\log \sum_{a'_t \in \mathcal{A} } p(X^n|a_t^{\prime n},h_t^n)$ penalizes all the actions.
%We emphasize that 
Both terms of Eq.~\ref{eq:overall} are critical for learning the generative language-grounded policy. When we train the model only with the language model term $\log p(X^n|a_t^{n},h_t^n)$ of Eq.~\ref{eq:overall},
the resulting neural network may not learn how to distinguish different actions rather than simply focusing on generating the instruction from the state observation.

For navigation, we use the model to capture the probability of the instruction conditioned on each action $a_t \in \mathcal{A}$.
%to obtain probabilities of generating the given instruction word sequence $X$ for all possible actions.
%We use $p(X|a_{t},h_t)$ to score the token sequence of the instruction.
The agent takes the action that maximizes the probability of generating the instruction:
%Then we take the argmax of those probabilities as
% \begin{equation}
$\arg\max_{a_t} p(X|a_{t},h_t)$.
% \label{eq:argmax}
% \end{equation}
%in time step $t$.
In other words, the language-conditional generative policy has a language model inside and navigates the environment by choosing an action that maximizes the probability of the entire instruction.
%This explains how the local movement can affect the global trajectory represented as the instruction.

% As an input to the neural network model, the next action candidate $a_t$ is computed from the directions to move next. This is widely allied in 
% We follow the methodoloy used in \citep{speaker-follower} ton compute the emdeddings of $a_t$. See Sec.~\ref{sec:nnmodel} for details.

%Experiment: In this paper,
% Later in the paper we empirically compare the discriminative and generative approaches, and show that the generative approach outperforms the discriminative approach in the unseen set in the vision-and-language navigation task.
%and generative approaches results in the best performance in the visual language navigation task.

\section{Related Work}

While most of previous studies~\citep{mattersim,vln-sm,vln-rcm,vln-areyoulooking,vln-press,prevalent}
%bases on the predictions of actions given the initial instructions along with past actions and scenes as of
have relied on the discriminative approach $p(a_{t}|X,h_t)$,
a few of previous studies~\citep{speaker-follower,vln-envdrop,cvpr2019_fast} have proposed the so-called speaker model which scores the instruction against the entire trajectory.
%the entire trajectories.
%In previous studies, however, 
Such speaker models are mainly used for two purposes;
(i) data augmentation with automatically generated trajectories~\citep{speaker-follower,vln-envdrop} and (ii) reranking the complete trajectories in beam decoding~\citep{speaker-follower,vln-envdrop,cvpr2019_fast}.
They however have not been used
%but not
%In these studies, the speaker models are trained for the candidates of the entire trajectories and never used for
for selecting local actions directly in either training or decoding.
%Precisely speaking, \citep{cvpr2019_fast} exploit the speaker function for 
%Some of recent studies utilize the language model loss as a part of pretraining or multi-task learning. Even in these studies, the generative language-grounded policy is not utilized to predict actions.
To the best of our knowledge, this paper is the first work that propose a standalone generative language-grounded policy for vision-and-language-navigation, that does {\it not} need the full state-action sequence nor to look ahead into the next state, before taking the action at each step. 
% and hence a language model function for predicting the next actions directly during navigation of the VLN task.

%\citep{vln-press} proposed a model with a discriminative approach that make use of pretrained language model and multiple instruction inputs for a single trajectory. 
%as of Table.2 in their paper.
%However multiple instructions are not always available in realistic navigation problem.% in because

Inspired by the the success of the embodied navigation datasets~\citep{roomnav,Matterport3D,touchdown}, new experimental settings and navigation tasks in realistic 3D modeling have been proposed,
such as
%embodied question answering~\citep{embodiedqa,eqa_matterport},
%expansions of the R2R dataset \citep{r4r,babywalk}
%and
dialog-based navigation tasks which include
vision-and-dialog navigation~\citep{cvdn},
vision-based navigation withlanguage-based assistanc~\citep{vnla},
and HANNA~\citep{hanna}.
Embodied question answering~\citep{embodiedqa,eqa_matterport},
interactive visual question answering~\citep{iqa} and ALFRED~\citep{alfred} for the navigation and object interaction are quite interesting task variants.
The proposed generative language-grounded policy is 
% directly 
applicable to these tasks where an agent solves a problem by following an instruction or having a conversation with another agent.

\section{Experimental Settings}

%\subsection{R2R Navigation Task}
\subsection{Datasets}

We conduct our experiments on the R2R navigation task~\citep{mattersim}, which is widely used for evaluating language-grounded navigation models and R4R~\citep{r4r}, which consists of longer and more complex paths when compared to R2R.
%is a dataset of VLN
%The R2R dataset is constructed on 3D house modelings of Matterport 3D~\cite{Matterport3D}.
R2R contains four splits of data: train, validation-seen, validation-unseen and test-unseen.
From the 90 scenes of Matterport 3D modelings~\citep{Matterport3D}, 61 scenes are pooled together and used as seen environments in both the training and validation-seen sets. Among the remaining scenes, 11 scenes form the validation-unseen set and 18 scenes the test-unseen set.
This setup tests the agent's ability to navigate in unseen environments in the test phase.
%in its first-look following an instruction
Some of previous studies make use of augmented datasets \citep{speaker-follower,vln-sm,vln-envdrop,cvpr2019_fast} in R2R experiments.
%Although the richer augmented datasets boost both the generative and discriminative models, 
We use the same augmented dataset from \cite{speaker-follower} which has been used in recent studies~\citep{vln-sm,cvpr2019_fast} for comparison.
% They are correspond to some paths in Matterport 3D modeling.

R4R was created based on R2R. In R4R, paths are composed of two paths drawn from R2R, implying that each reference path in R4R is not necessarily the shortest path between the starting point and the goal point.
R4R is more suitable for evaluating how closely the agent follows a given instruction that corresponds to a long and complex path. 
R4R consists of train, validation-seen and validation-unseen sets, but does not contain the test-unseen set, unlike R2R. 
We provide more detailed statistics of R2R and R4R in Appendix~\ref{app:dataset}.

%R2R require the generalization abilities of models in unseen conditions.
%Following \cite{spl}, we compare our models in the realistic,
%\subsection{Experimental settings}
%\subsection{Experiment Condition}

%\subsection{Neural Network and Training}
\subsection{Neural Network Models}
\label{sec:nnmodel}

%\subsection{Explainer}
%We train both our proposed explainer and follower of \citep{speaker-follower} with BERT. We evaluate the model of combination of explainer and follower.

%\subsection{Neural Network}

We use the network architecture of the speaker from \citep{speaker-follower} to implement generative policies which include a language model $p(X|a_t,h_t)$. 
%This language model is pretrained in the same way as \citet{speaker-follower}.
% Their speaker neural network is used for the data augmentation and reranking for beam decoding results, and is not used in greedy navigation trials in \citet{speaker-follower}.
We also use the follower network architecture by \citet{speaker-follower} for implementing discriminative policies.
%We use the network architectures of the speaker and follower models from \citet{speaker-follower} to implement our generative and discriminative models.
%For generative language-grounded policy learning, we use the same network used by \cite{speaker-follower}.
%, which are referred as ``speaker model''.
%We also train the same follower model $p_f(a_t|X,h_{t})$ from their work as the discriminative policy.
%Both the speaker and follower model of \cite{speaker-follower} exploits the possible directions to move next to encode the embeddings of the next actions.
We follow \citet{speaker-follower} and create the embedding of the next action by concatenating the 4-dimensional orientation feature $[\sin\phi; \cos\phi; \sin\theta; \cos\theta]$ and the image feature extracted from a pretrained ResNet~\citep{resnet}, where $\phi$ and $\theta$ are the heading and elevation angles, respectively.
%Our speaker model scores the given instruction at each time step with these embeddings of the next actions that are also used in the follower model of \cite{speaker-follower}. Our speaker model is strictly prohibited to ``look-ahead'' the next state before the actions.
Both generative and discriminative models use the panoramic view and action embedding, following \cite{speaker-follower}.
The generative policy scores an instruction based on the embedding of each of the next possible actions and the state representation which is also used by the discriminative policy.

\paragraph{Navigation}

%In some of the previous studies, beam decoding approaches have been shown to achieve high task success rates while they result in very low, $O(0.01)$ of SPL because of their unrealistically long transition length in total navigation trials.
In all our experiments, a single agent navigates in an environment only once given a single instruction, for each task, because it is unrealistic to have multiple agents simultaneously navigating in an indoor, hosehold environment. This implies that we do not use beam search nor pre-exploration in unseen environments. See \citet{spl} for more discussion on the condition and evaluation of the navigation task.

\subsection{Training}

\paragraph{R2R} 

We first train a language model that predict an instruction from the entire trajectory
in the same way as \citet{speaker-follower} from the dataset. We finetune each policy using imitation learning, where we let the policy navigate the environment and give the action that leads to the shortest path at each time step as supervision,  closely following \citet{mattersim}. Just like \citet{speaker-follower}, we start training a policy with both the augmented and original training sets, and then switches to using the original training set alone.

\paragraph{R4R} 

we first train a language model for the generative policy from the R4R dataset.
Since there are more than 10 times more training instances in R4R than in R2R, we do not augment data.
Unlike in R2R, we test both learning strategies; supervised learning and fidelity-oriented learning. 
In the case of supervised learning, we train both our generative and discriminative policies to maximize the log-probability of the correct action from the training set~\citep{speaker-follower}. On the other hand, fidelity-oriented learning is a particular instantiation of imitation learning, in which a set of heuristics are used to determine the correct next action based on the proximity of the current state to the reference trajectory at each time step. We describe fidelity-oriented learning in Appendix~\ref{sec:r4r_fidelity}, and also summarize any remaining training details in Appendix~\ref{sec:sampling}.

\begin{table*}[t]
\begin{center}
	\scriptsize\begin{tabular}{lp{1.4em}p{1.2em}p{1.4em}p{1.4em}p{1.4em}p{1.6em}c |p{1.4em}p{1.2em}p{1.4em}p{1.4em}p{1.4em}p{1.6em}c}
        \toprule
        & \multicolumn{7}{c}{Validation (Seen)} & \multicolumn{7}{c}{Validation (Unseen)}  \\
Model                  & ~~PL$\downarrow$ & ~NE$\downarrow$ & ~~SR$\uparrow$ & SPL$\uparrow$ & CLS$\uparrow$ & nDTW$\uparrow$ & ~~SDTW$\uparrow$\hspace{-0.5em} & ~~PL$\downarrow$ & ~NE$\downarrow$ & ~~SR$\uparrow$ & SPL$\uparrow$ & CLS$\uparrow$ & nDTW$\uparrow$ & ~~SDTW$\uparrow$\hspace{-0.9em} \\
\midrule
Disc.                                        & 10.69 & 5.40 & 0.519 & 0.482 & 0.619 & ~0.588 & 0.445 & 12.88 & 6.52 & 0.380 & 0.335 & 0.488 & ~0.458 & 0.304 \\
Disc. +Aug.\hspace{0.05em}(A) \hspace{-3em}  & 10.60 & 5.15 & 0.525 & 0.489 & 0.633 & ~0.596 & 0.445 & 12.05 & 6.22 & 0.431 & 0.392 & 0.528 & ~0.496 & 0.356  \\
Gen.                                         & 11.23 & 5.53 & 0.481 & 0.451 & 0.625 & ~0.579 & 0.427 & 12.98 & 6.17 & 0.434 & 0.371 & 0.514 & ~0.478 & 0.344 \\
Gen.  +Aug. (B) \hspace{-3em}                & 11.45 & 4.78 & \textbf{0.563}  & \textbf{0.531} & \textbf{0.664} & ~\textbf{0.630} & \textbf{0.505} & 13.92 & 4.78 & \textbf{0.476} & \textbf{0.405} & \textbf{0.539} & ~\textbf{0.503} & \textbf{0.379} \\
\midrule
Gen.+Disc.(A+B)\hspace{-0.5em}               & 10.18 & 4.67 & 0.568 & 0.540 & 0.680 & ~0.640 & 0.510 & 12.06 & 5.42 & 0.489 & 0.437 & 0.570 & ~0.533 & 0.403 \\
Gen.+Disc.(A+B)$^*$\hspace{-1.5em}           & 11.30 & 4.58 & 0.575 & 0.541 & 0.678 & ~0.636 & 0.509 & 14.65 & 5.19 & 0.518 & 0.439 & 0.564 & ~0.515 & 0.397 \\
\bottomrule
	\end{tabular}
    \caption{
    Performance of \textit{generative policies} and \textit{discriminative  policies} on the R2R dataset.
    %Comparisons of \textit{generative language-grounded policy} and \textit{discriminative policy} of the follower model in \citep{speaker-follower}.
    +Aug. represents policies trained with the augmented dataset by \citep{speaker-follower}.
    Bold fonts are used for the best result
    as a single model in major metrics. $*$ represents the use of backtracking of FAST.
    %Note that we apply stochastic sampling of Sec.~\ref{sec:sampling} and therefore the baseline results of discriminative policy are better than reported in \citep{speaker-follower}.
    }
    %\vspace{-0.5em}
    \label{table:speakervsfollower}
\end{center}

\vspace{-4mm}
\end{table*}

\subsection{Evaluation Metrics}

We use the following four metrics that have been commonly used to assess a policy in R2R: path length (PL) of the entire trajectory, navigation error (NE), success rate (SR) and success weighted by path length (SPL).
%Since all transitions of agents need to be considered, beam models require a large amounts of trajectories in total.
%\end{description}
Among those evaluation metrics, we consider \textbf{SR} and \textbf{SPL} as primary ones for R2R because they are derived from the number of successes trials in the navigation.
%In the case of R4R, 
We report CLS~\citep{r4r}, nDTW and SDTW~\citep{SDTW} in addition to the four metrics above, as these additional metrics are better suited to longer and more complex paths. 
These three metrics are based on the distance between the policy's trajectory and the reference path. 
Following \citet{babywalk}, we use \textbf{CLS} and \textbf{SDTW} as primary metrics for R4R.
We avoid using SPL for the R4R evaluation because it is not suitable for R4R performance comparisons~\citep{r4r}.
See Appendix~\ref{sec:metrics} for the detailed description of each metric.

%\subsection{Baselines}

\begin{table*}[t]
\begin{center}
	\scriptsize\begin{tabular}{lcccc|cccc|cccc}
        \toprule
                 & \multicolumn{4}{c}{Validation (Seen)} & \multicolumn{4}{c}{Validation (Unseen)} & \multicolumn{4}{c}{Test (Unseen)} \\
Model            & PL$\downarrow$ & NE$\downarrow$ & SR$\uparrow$ & SPL$\uparrow$ & PL$\downarrow$ & NE$\downarrow$ & SR$\uparrow$ & SPL$\uparrow$ & PL$\downarrow$ & NE$\downarrow$ & SR$\uparrow$ & SPL$\uparrow$ \\
\midrule
Random           &  9.58 & 9.45 & 0.16 &    - &  9.77 & 9.23 & 0.16 &    - &  9.93 & 9.77 & 0.13 & 0.12 \\
Seq2seq          & 11.33 & 6.01 & 0.39 &    - &  8.39 & 7.81 & 0.22 &    - &  8.13 & 7.85 & 0.20 & 0.18 \\
RPA              &     - & 5.56 & 0.43 &    - &     - & 7.65 & 0.25 &    - &  9.15 & 7.53 & 0.25 & 0.23 \\
Speaker-Follower &     - & 3.36 & 0.66 &    - &     - & 6.62 & 0.35 &    - & 14.82 & 6.62 & 0.35 & 0.28 \\
Self-Monitoring  &     - &    - &    - &    - &     - &    - &    - &    - & 18.04 &	5.67   & 0.48 & 0.35 \\
RCM+SIL (train)  & 10.65 & 3.53 & 0.75 & 0.67 & 11.46 & 6.09 & 0.50 & 0.42 & 11.97 & 6.12 & 0.43 & 0.38 \\
EnvDrop          & 11.0  & 3.99 & 0.62 & 0.59 & 10.70 & 5.22 & 0.52 & 0.48 & 11.66 & 5.23 & 0.51 & \textbf{0.47} \\
FAST             &     - &    - &    - &    - & 21.17 & 4.97 & 0.56 & 0.43 & 22.08 & 5.14 & \textbf{0.54} & 0.41 \\
PRESS            & 10.57 & 4.39 & 0.58 & 0.55 & 10.36 & 5.28 & 0.49 & 0.45 & 10.77 & 5.49 & 0.49 & 0.45 \\
Gen.+Disc. Policy & 11.30 & 4.58 & 0.57 & 0.54 & 14.65 & 5.19 & 0.52 & 0.44 & 14.31 & 5.24 & \textbf{0.54} & \textbf{0.46} \\
\midrule
\midrule
Human            & -&-&-&- & -&-&-&- & 11.90 & 1.61 & 0.86 & 0.76 \\
\bottomrule
	\end{tabular}
    \caption{
    Comparison of baselines and the proposed policy under single run experimental setting on the R2R dataset.
    Bold fonts for the first and second best results in SR and SPL.% in the test set.
    }
    %\vspace{-0.5em}
    \label{table:overallresult}
\end{center}

\vspace{-4mm}
\end{table*}
%\midrule

\section{Results}

We use R2R as the main task to investigate the efficacy of and analyze the behaviour of the proposed language-grounded generative policy and its relative performance against existing approaches, as R2R has been studied more extensively than R4R has. We thus present the result on R2R first, and then finish the section with the result on R4R. 

\subsection{Generative vs. Discriminative Policies}
\label{sec:gen_disc}

Table~\ref{table:speakervsfollower} shows the performances of the generative language-grounded policy (Generative Policy) and discriminative policy (Discriminative Policy) in the R2R dataset. We show the result with and without data augmentation.
All the policies were trained with a stochastic mixture of supervised learning and imitation learning, resulting in better performance than those reported by \cite{speaker-follower}, even in the case of the discriminative baseline.

The first observation we make is that data augmentation has a bigger effect on the validation-unseen split than on the validation-seen split. This suggests that the main effect of data augmentation is to make a policy more robust to the changes in environments by discouraging the policy from overfitting to environments that are seen during training. This effect is observed with both discriminative and generative policies.
However, we consider the discriminative policies are easy to overfit to seen environments in the training time especially without the augmentated dataset.
In the validation-unseen split, the generative policy always performs better than the discriminative one in both SR and SPL.

Second, when data augmentation was used, the proposed generative policy outperforms the discriminative policy in both validation-seen and validation-unseen splits. This is particularly true with the primary metrics, SR and SPL. The path length (PL) is generally longer with the generative policy, but the difference is within 1 meter on average. 

Finally, the best performing approach is the combination of the discriminative and generative policies (both trained with data augmentation). This clearly indicates that these two approaches are capturing two different aspects of visual language navigation. Back-tracking further improves this hybrid policy in terms of SR, although the improvement in SPL is minimal, as back-tracking introduces extra transitions. 

In CLS, nDTW and SDTW, the generative policy achieves higher performance than the discriminative policy does, which suggests that the proposed generative policy follows the reference path more closely compared to the discriminative one. We conjecture this is because the generative policy is sensitive to the language instructions by construction.

% We present CLS, nDTW and SDTW scores in the R2R seen and unseen validation in Appendix~\ref{sec:r2r_cls}.
% Table~\ref{table:sdtw}

%The Discriminative Policy achieves better performance than the Generative Policy without the augmentation in the seen validation split.
%Especially he generative language-grounded model achieves the best result in the validation unseen datasets.
% In the validation-seen split, the generative policy trained with the augmented dataset achieves the better results than the discriminative counterpart. This however did not hold without augmentation, suggesting that generative policies may require more data than discriminative policies do.

%We also notice t

%We confirm that the scores speaker-driven exploration is better than those of follower.
%Especially, LS is better than the follower in SR in a great margin.

% We also notice that the combination of the generative and discriminative policy achieves the best result.
%The FAST proposed the back-tracking framework to choose the goal place from visited places for the single agent.

% than those in SR.

\subsection{Comparison against Baselines}

Table~\ref{table:overallresult} lists the performances in the validation-seen, validation-unseen and test-unseen sets in R2R, collected from the public leaderboard and publications. We achieve near state-of-the-art result only with the original training set and augmented dataset released by \cite{speaker-follower}.
We compare our approach against the following previous baselines:
Random~\citep{mattersim}, Seq2Seq~\citep{mattersim}, RPA~\citep{vln-rpa}, Follower~\citep{speaker-follower}, Self-Monitoring~\citep{vln-sm}, RCM~\citep{vln-rcm}, EnvDrop~\citep{vln-envdrop}, FAST~\citep{cvpr2019_fast} and PRESS~\citep{vln-press}. They are described in detail in Appendix~\ref{sec:baselines}.
All of them, except for the random agent, follow the discriminative approach, unlike our proposal. 

%We also notice that some of previous models such as EnvDrop, FAST and PRESS tend to result in  scores in either SR or SPL metric. Our model achieve best or nearly best result among baseline models.
In terms of SR, our model performs comparably to FAST which uses the same neural network by \citet{speaker-follower}, while our model is better in SPL.
In terms of SPL, our model is the second best only next to the EnvDrop model.\footnote{
Our reported SPL is 0.4647 only marginally lower than 0.47 of EnvDrop.
}
Our policy however ends up with a better
%than the EnvDrop model
SR than EnvDrop does.
Overall, the proposed approach is equivalent to or close to the existing state-of-the-art models in both SR and SPL.

%We also  the result from the recenPLy-proposed PREVALENT model \cite{prevalent}.
%PREVALENT 
The recently proposed PREVALENT model \citep{prevalent}
benefits from large scale cross-modal attention-based pretraining.
They apply extensive data augmentation to create 6,482K image-text-action triples for pretraining, unlike the other approaches in Table~\ref{table:overallresult}.
% with eight V100 GPUs.
Thanks to this extensive augmentation, they achieve SR of 0.54 and SPL of 0.51.
On the other hand, we only use 178K augmented examples from \cite{speaker-follower}, widely used in previous studies~\citep{vln-sm,cvpr2019_fast}, for more direct comparison with previous studies.
Although we have nevertheless achieved the comparable SR with an order of magnitude smaller augmented data, we expect our approach would further improve with this more aggressive augmentation strategy in the future. 
% against PREVALENT.
%\textcolor{red}{We also note that we achieve this without any pretrained language encoders such as BERT. We discuss further for the effect of pretrained language model on VLN in the Appendix}~\ref{sec:bert}.

\begin{figure*}[t]
\begin{minipage}{0.5\hsize}
\begin{center}
\vspace{-1em}
\includegraphics[scale=0.45,clip]{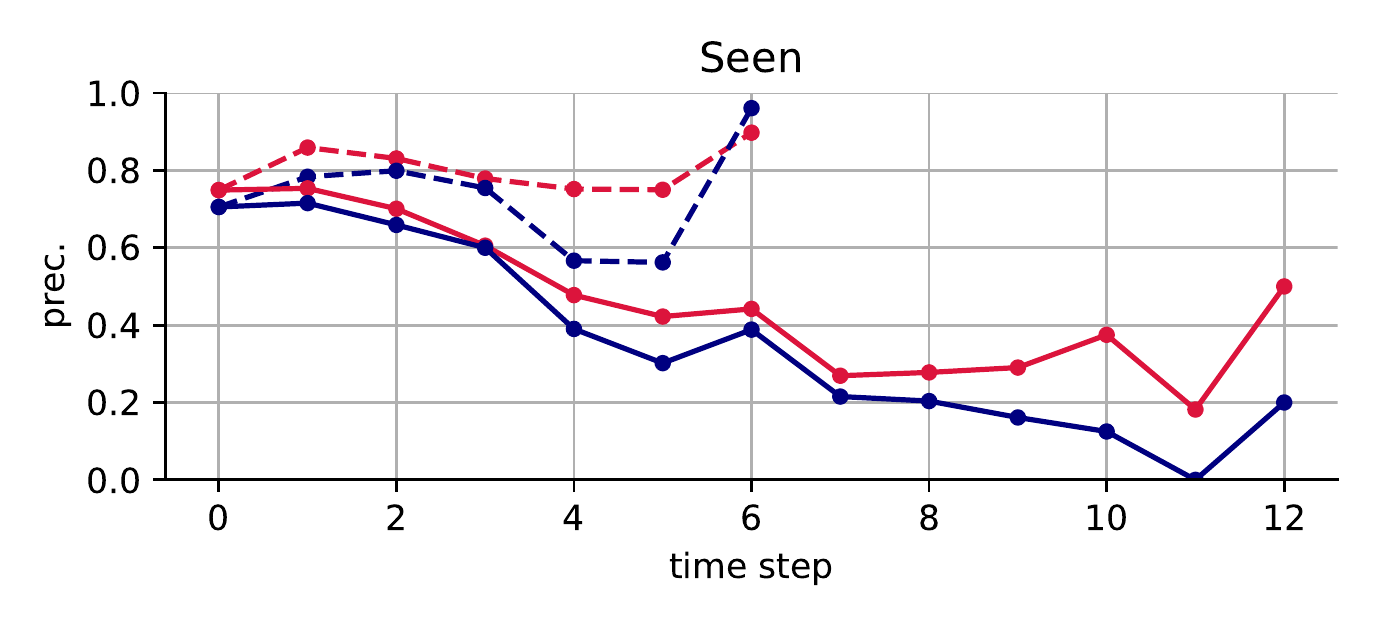}
\end{center}
\end{minipage}
\begin{minipage}{0.5\hsize}
\begin{center}
\vspace{-1em}
\hspace{-2em}
\includegraphics[scale=0.45]{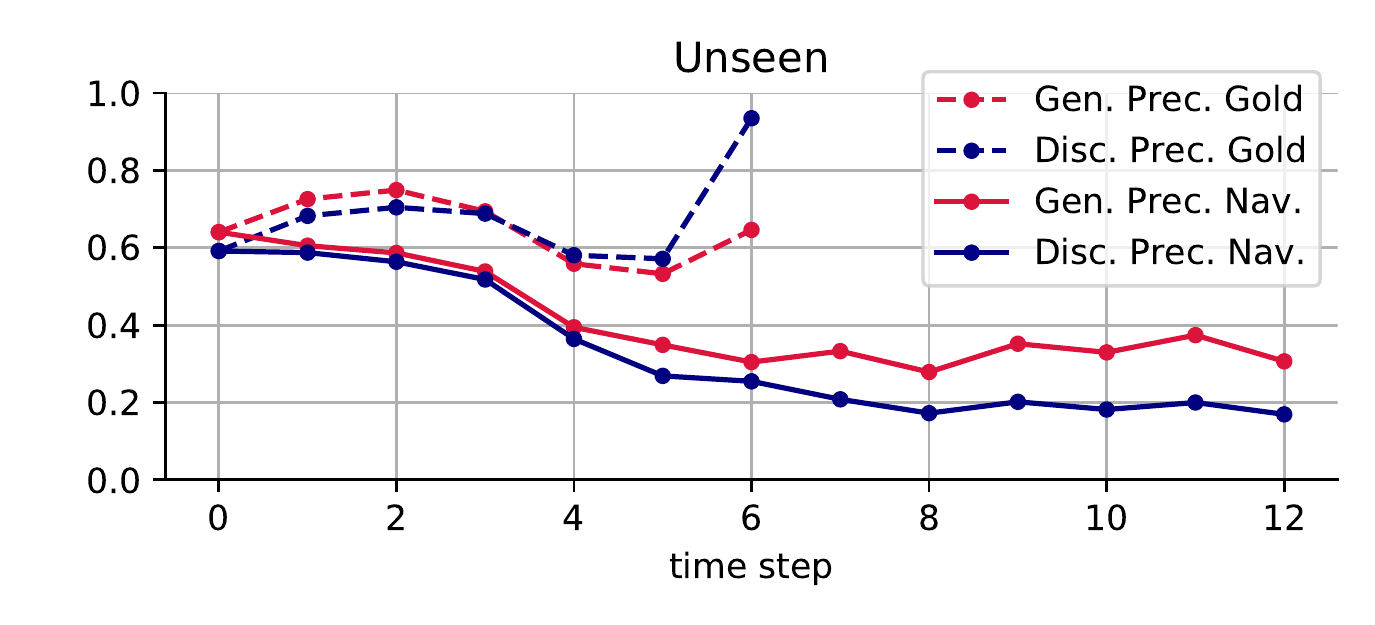}
\end{center}
\end{minipage}
\vspace{-1em}
%%%%%%%%%
% \begin{minipage}{0.5\hsize}
% \begin{center}
% \vspace{-1em}
% \includegraphics[scale=0.45,clip]{fig/agreement_seen_pub.pdf}
% \end{center}
% \end{minipage}
% \begin{minipage}{0.5\hsize}
% \begin{center}
% \vspace{-1em}
% \hspace{-2em}
% \includegraphics[scale=0.45,clip]{fig/agreement_unseen_pub.pdf}
% \end{center}
% \end{minipage}
% \vspace{-1em}
\caption{
The precision of actions by the generative (red) and discriminative (blue) models on the reference trajectory (dashed lines) and on navigation trajectories (solid lines).
%Solid lines are accuracy for all actions. Dotted lines are accuracy of actions when the correct actions are ``STOP'' at that place, while the dashed lines are accuracy of actions when the correct actions are not ``STOP'' at that place.
% \textbf{Bottom}: the agreement of actions between the generative and discriminative models  on shortest paths (dashed lines) and on navigation trials (solid lines). The horizontal axis corresponds to the time step of trials.
}
\label{fig:prec_action}
\end{figure*}

% \begin{figure*}[t]
% 	\hspace{0em}
%     \begin{center}
% 	\includegraphics[scale=0.55,clip]{fig/bleu_graph0.pdf}
% 	\vspace{0em}
%       \caption{
% 			 BLEU scores for each time step $t$ given the gold instructions.
% 			 }
%       \label{fig:bleu}
%     \vspace{0em}
%     \end{center}
% \end{figure*}

% \subsubsection{Instruction Generation from Language-Grounded Models}
% Figure~\ref{fig:bleu} show the relation of BLUE and languge modrl coefficient $\lambda$ of  Eq.~\ref{eq:lmc}.

% \subsection{Further Analyses}

\subsection{Action prediction accuracy}
\label{sec:analysis_comp}

Figure~\ref{fig:prec_action} plots the precision of predicted actions over time on the validation-seen and validation-unseen sets in R2R for both the generative policy and the discriminative language-ground policies.
We use the discriminative and generative policies notated as A and B in Table~\ref{table:speakervsfollower} for this analysis.
%and examine the prediction of actions from generative and discriminative policies respectively.
%The solid lines represent the precision of actions for Gen 
%The precision and  at time step $t$.
When the agents are presented with the gold trajectories,
%of the shortest path,
%as the dashed lines,
both policies predict actions more accurately than they would with their own trajectories.
%real navigation.
%Obviously this is not true in navigation trials and 
In real navigation, the action selection error accumulates, and prediction by both policies degrades over time.
%become inaccurate as the solid lines.
The generative policy, however, is more tolerant to such accumulated error than the discriminative policy is, achieving a higher precision in later steps.
This is especially the case in unseen environments. Additional analyses for the difference of policies are in Appendix~\ref{app:polices_diff}.

%The precision of Disc model drops greater than those of Gen in navigation trials, while the Gen model retains a certain level of accuracy even in later time steps.

\begin{comment}
To visualize the precision of ``STOP'' prediction, we draw additional lines: dashed lines for the precision of actions where an agent should not choose ``STOP'' such as at the beginning of the navigation and dotted lines for precision of actions where an agent should to stop there.
We notice that the the blue dashed lines of precision of Gen. are always above the red dashed lines of Disc \footnote{The dashed lines overlaps the solid lines at small $t$ because of the lacks of the gold ``STOP'' actions at the beginning.
%because there are  no correct actions for ``STOP'' at the beginning of the navigation.
}. However the red dotted lines are often often the blue dotted lines.
This suggests that Gen model tends to continue the navigation while the Disc model outputs ``STOP''.
Overall, an agent with Gen model is effective from the beginning to the end of the navigation while Disc model is beneficial when the agent needs to choose the ``STOP'' action.
We present further analyses of predicted ``STOP'' actions with F1 measure in Appendix. In the analyses with F1, our Gen. model perform indeed better than Disc. model even with large time step $t$.
\end{comment}

\subsection{Token-wise Prediction Entropy}

The proposed generative policy allows us to easily inspect how it uses the instruction.
%We further analyze how the agent with generative language-ground policy utilizes the instruction during the navigation.
A few tokens in an instruction often have significant influence on the agent's decision. For example, if an instruction ends with ``...then stop at the kitchen'' and the agent is between the kitchen and dinning room, the token ``kitchen''
%has specific meaning to
decides when the agent predicts the ``STOP'' action.
Since the generative language-grounded policy relies on token-wise scoring of the instruction given each action,
we can directly measure how each token in the instruction affects the action prediction. We call this measure 
%we examine the relationship between each action and specific tokens in the instruction.
% The agent chooses the next action $a_t$ from the language model $p(X|a_t,h_t)$ as of Eq.~\ref{eq:argmax}.
% An instruction is a sequence of tokens $X=(w_0, w_1,...,w_k,...)$ and therefore
% $\log p(X|a_t,h_t)=\sum_k \log p(w_k|a_t,h_t,w_{:k-1})$.
% We introduce 
token-wise prediction entropy (TENT) and define it as 
\begin{equation}
S(w_k)=-\sum_{a_t \in \mathcal{A}}{q(a_t,w_k)\log_{|\mathcal{A}|} q(a_t,w_k)},
\end{equation}
where $S(w_k) \in \left[0, 1\right]$,
% $q(a_t,w_k)$ is the probability of the action $a_t \in \mathcal{A}$ and 
$\mathcal{A}$ is the action set, and
%normalized over the possible actions $\mathcal{A}$:
\begin{equation}
q(a_t,w_k) = \frac{p(w_k|a_t,h_t,w_{:k-1})}{\sum_{a_t \in \mathcal{A}} p(w_k|a_t,h_t,w_{:k-1})}.
\end{equation}
% TENT is easy to compute for the generative language-grounded policy because it already computes $\log p(w_k|a_t,h_t,w_{:k-1})$ for all possible actions $a_t$ during navigation.
When some tokens are influential to the action prediction, the entropy of scoring these tokens will be low. Otherwise, when $S(w_k)$ is close to 1, 
% which suggests that
%$S(w_t) \to 1$, which suggests that
%This is case, because
% $p(w_k|a_t,h_t,w_{:k-1})$ is almost same for any action, and that
% for the prediction of this token
%In this case,
the corresponding token $w_k$ is deemed less influential for the next action prediction.
%We visualize how each token affects the agent decision with TENT in Figure~\ref{fig:TENT_exam}.
% TENT always satisfies $0 \leq S(w_k) \leq 1$, and we 
We visualize $1-S(w_k)$, to which we refer as 1-TENT,  
% in visualization 
to identify highly influential tokens at each time step.
% illustrate positive peaks for the large influence in the action prediction.
% We refer to $1-S(w_k)$ as 1-TENT.

\begin{figure*}[t]
\begin{center}
\includegraphics[scale=0.40,clip]{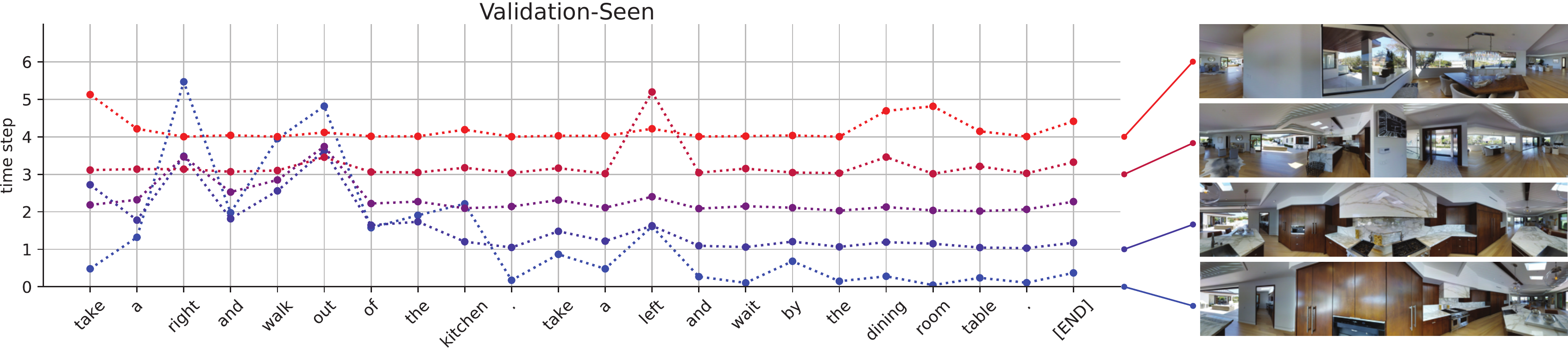}
\end{center}
\vspace{-1em}
\begin{center}
\includegraphics[scale=0.40,clip]{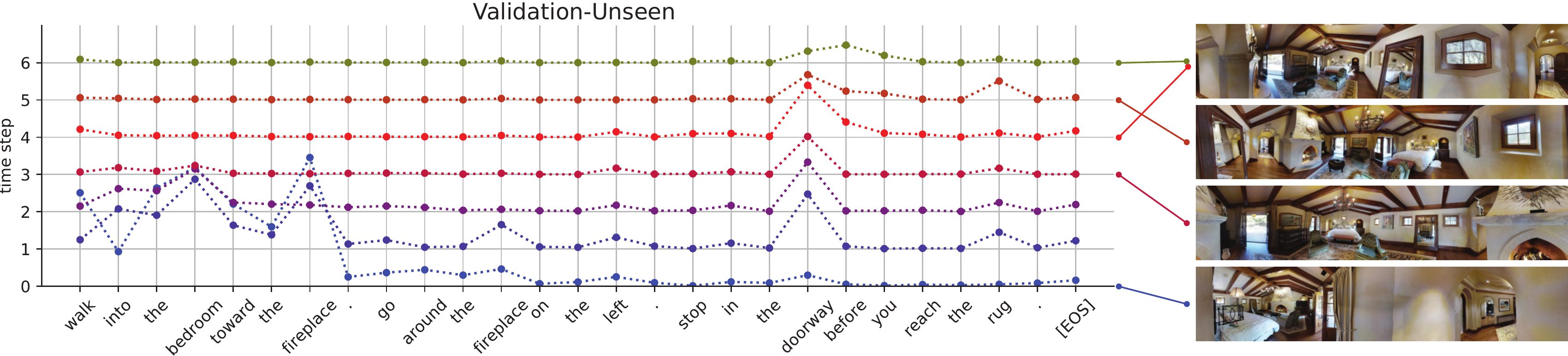}
\end{center}
\vspace{-1em}
\caption{Token-wise prediction entropy (TENT) for two navigation instances from validation-seen (top) and validation-unseen (bottom) sets in R2R.
%Horizontal axis is the position of each token in the instructions.
% Tokens of the instruction are aligned on the horizontal axis.
%Vertical-axis corresponds the TENT by each time step.
The vertical axis corresponds to the 1-TENT drawn at each time step $t \in \mathbb{N} \cup \{0\}$, as $t+\frac{1}{\Delta} (1-S(w_k))$, where $\Delta=0.05$ so that
%is the scale of 1-TENT in one vertical-tick in the graphs.
one vertical-tick corresponds to 0.05.
% in the scale of 1-TENT.
We draw multiple lines that correspond to different time steps colored from blue to red and green.
We attach the panoramic views for some of the trial time steps.
%Here we can observe five isolated downward peaks.
%They corresponds to words ``hallway'', ``kitchen'', ``bathroom'', ``far'' and ``gym'' in the instruction.
}
\label{fig:TENT_exam}
\end{figure*}

Figure~\ref{fig:TENT_exam} visualizes how actions are related to each token in each time step with two sample navigation trajectories from the validation-seen and validation-unseen splits in R2R.
We use the generative policy trained with data augmentation from Table~\ref{table:speakervsfollower}. 
% without back-tracking. 
Both trials end successfully within five and seven time steps, and we plot five and seven curves of 1-TENT.
In the early stage of the navigation ($t<3$),
%We firstly notice that for small time steps,
initial tokens exhibit large 1-TENT, meaning the change of actions yields a great difference in those token predictions.
This tendency is observed in both seen and unseen environments.
%This suggests that the difference of the initial actions results in great differences of the instruction scoring.
We conjecture
%that the prediction greatly differs action by action
this is a natural strategy learned by the policy to take when there is no navigation history.
%We then notice
%In the later stage, the agent often shows high 1-TENT for specific tokens.
%When the policy lacks the history, the action prediction relies on the early part of the instruction.

In the seen navigation example, the agent is asked to navigate from the kitchen to the dinning room table. In the initial steps, the agent tries to go out from the kitchen, and phrases such as ``right'' and ``walk out'' have high 1-TENT. At $t=3$, the agent is out of the kitchen and needs to turn left at the middle of the large room with high 1-TENT on ``left''.
% (see the panorama view on the right). 
Finally, the agent finds the dinning table and stops there with the high 1-TENT for the tokens indicating the stop point.

In the unseen navigation instance, the agent is asked to navigate from the hallway, cross the large bedroom and stop outside the carpet.
In the trial, the agent first moves toward the goal node based on the keywords ``bedroom'' and ``fireplace''.
%Then it seems that
It also exhibits high 1-TENT for ``doorway'', which is a clue for identifying the goal node.
%in order to find out the stop place.
%We here notice that
%The agent then
This agent, however, passes the node of the success for the first time at $t=4$.
At $t=5$, the agent has the high 1-TENT for both ``doorway'' and ``rag'' and then goes back to the same place with $t=4$.
Finally, it stops with the high 1-TENT for ``before'' and the slight 1-TENT for ``rag'' at $t=6$.
%, as if it is determining the last node to stop.
As we have seen here, the agent has different 1-TENT values depending on the context even if it is in the same place.

%As we see here, token-wise entropy prediction and 1-TENT reflect why the agent takes a specific action by showing the variety of each token prediction in the reference instruction.
%We also notice that the agent sometimes has high 1-TENT for tokens difficult to interpret such as the initial tokens of instructions at the last step. Therefore, we need to carefully examine this.
%However,
% 1-TENT is computed from $\log p(X|a_t,h_t)=\sum_k \log p(w_k|a_t,h_t,w_{:k-1})$, which is directly used for the next action prediction in Eq.~\ref{eq:argmax}.
% with summation over tokens.
%Therefore 1-TENT is a measure easy to interpret why it affects the action prediction.
% in the proposed generative language-grounded policy.
Although the result of the 1-TENT visualization is similar to the attention maps \citep{bahdanau2014neural,transformer}, 1-TENT is much more directly related to the action prediction.
The attention map represents the internal state of the neural network, while 1-TENT is based on the output of the neural network.
This property makes the proposed 1-TENT a powerful tool for investigating and understanding the generative language-grounded policy.

%\footnoate{Since panoramic view sight of \cite{speaker-follower} include panoramatic view of elevation and tilt angle of }.
%because the agent is able to learn the combination of specific words and visual scenes
%Therefore we assume 1-TENT is a novel for the interpretability of the model behaviors in the generative language-grounded policy agent.
%We also notice that we are able to apply the trained neural network for generating a textual sequence given an action. However we made a prudent decision not to adapt this because our network is not trained for this and our neural network is not used for generating some textual sequences

%\subsection{Instruction Prediction}

%the model performs better than the model with discriminative (Disc) in terms of the action prediction when the correct actions are not ``STOP'' as of the dashed lines in the figures. Note that there are no gold actions for ``STOP'' at the beginning of the navigation and hence the dashed lines overlaps the solid lines. 

\subsection{R4R}

% we use reward1

%SPL is not adequate for R4R because the reference path is not the shortest path.

\begin{table*}[t]
\begin{center}

\scriptsize\begin{tabular}{lp{1.4em}p{1.2em}p{1.4em}p{1.4em}p{1.6em}c|p{1.4em}p{1.2em}p{1.4em}p{1.4em}p{1.6em}c}
        \toprule
        & \multicolumn{6}{c}{Validation (Seen)} & \multicolumn{6}{c}{Validation (Unseen)}  \\
Model                  & ~~PL$\downarrow$ & ~NE$\downarrow$ & ~~SR$\uparrow$ & CLS$\uparrow$ & nDTW$\uparrow$ & ~~SDTW$\uparrow$\hspace{-0.5em} & ~~PL$\downarrow$ & ~NE$\downarrow$ & ~~SR$\uparrow$ & CLS$\uparrow$ & nDTW$\uparrow$ & ~~SDTW$\uparrow$\hspace{-0.9em} \\
\midrule
RCM~~~~fidelity-oriented                & 18.8 & 5.4 & 0.526  & 0.553 & ~- & - & 28.5 & 5.4  & 0.261 & 0.346 & - & - \\
nDTW~~fidelity-oriented            &-&-&-&-&~-&-& - & - & 0.285 & 0.354 & 0.304 & 0.126 \\
BabyWalk~~IL+RL      &-&-&-&-&~-&-& 22.8 & 8.6  & 0.250 & 0.455 & 0.344 & 0.136 \\
BabyWalk~~IL+RL+Cur. &-&-&-&-&~-&-& 23.8 & 7.9  & 0.296 & 0.478 & 0.381 & \textbf{0.181} \\
\midrule
Disc.~~supervised         & 20.1 & 7.0 & 0.386 & 0.622 & 0.512 & 0.305 & 20.0 & 9.8  & 0.172 & 0.446 & 0.305 & 0.101 \\
Disc.~~fidelity-oriented  & 21.1 & 6.6 & 0.449 & 0.644 & 0.530 & 0.360 & 29.2 & 9.2  & 0.211 & 0.385 & 0.282 & 0.116 \\
Gen.~~~supervised          & 19.8 & 8.8 & 0.316 & 0.563 & 0.442 & 0.246 & 19.7 & 9.8  & 0.193 & \textbf{0.479} & 0.325 & 0.121 \\
Gen.~~~fidelity-oriented   & 21.0 & 6.9 & 0.448 & 0.629 & 0.517 & 0.349 & 22.8 & 8.7  & 0.255 & 0.471 & 0.348 & 0.162 \\
\bottomrule
	\end{tabular}
	
    \caption{
    Performance of \textbf{gen}erative and \textbf{disc}riminative  policies on the R4R validation-seen and -unseen splits.
    We use bold-face to indicate the best models according to CLS and SDTW in the validation-unseen split.
    % ``BabyWalk -curriculum`` is the BabyWalk agent without curriculum-based learning.
    }
    %\vspace{-0.5em}
    \label{table:r4r}
\end{center}

\vspace{-8mm}
\end{table*}

% \section{Results of R4R Seen Validation Set}
% \label{sec:r4r_seen}

% \begin{table*}[t]
% \begin{center}
% 	\scriptsize\begin{tabular}{lcccccc}
%         \toprule
%         & \multicolumn{6}{c}{Validation (Seen)}  \\
% Model   & PL$\downarrow$ & SR$\uparrow$ & SPL$\uparrow$ & CLS$\uparrow$ & nDTW$\uparrow$ & SDTW$\uparrow$  \\
% \midrule
% Discriminative Policy teacher-forcing   & 20.1 & 0.386 & 0.364 & 0.622 & 0.512 & 0.305 \\
% Discriminative Policy fidelity-oriented & 21.1 & 0.449 & 0.421 & 0.644 & 0.530 & 0.360 \\
% Generative Policy teacher-forcing       & 19.8 & 0.316 & 0.302 & 0.563 & 0.442 & 0.246 \\
% Generative Policy fidelity-oriented     & 21.0 & 0.448 & 0.416 & 0.629 & 0.517 & 0.349 \\
% \bottomrule
% 	\end{tabular}
%     \caption{
%     Performance of \textit{generative policies} and \textit{discriminative  policies} on the R4R seen validation set.
%     Bold fonts are used for the best models in CLS and SDTW.
%     }
%     %\vspace{-0.5em}
%     \label{table:r4r_seen}
% \end{center}
% \end{table*}

Table~\ref{table:r4r} presents the results on R4R with baseline model performance.\footnote{
We describe the baseline models in Appendix~\ref{sec:baselines}.
}
Similarly to our earlier observation on R2R, the proposed generative policy works as well as or better than the discriminative policy as well as the other baselines, in terms of the primary metrics which are CLS and SDTW in this case especially in the validation-unseen split. The generative policy trained with supervised learning outperforms all the baseline policies in CLS, while the generative policy trained with imitation learning is close to BabyWalk trained with reinforcement learning (IL+RL) and curriculum learning (IL+RL+Cur.) in SDTW. As both reinforcement learning and curriculum learning could also be applied to our approach, we expect to this gap to completely close in the future. 

% When we compare the result of the discriminative and generative policies on the R4R unseen validation split, we again see a similar trend. Regardless of the choice of a learning algorithm, the generative policies outperform the discriminative ones across all the metrics. 

As we have observed on R2R, without data augmentation, the discriminative policy works as good as or often better than the generative approach does on R4R. the generative policy is however significantly better than the discriminative one in the validation-unseen split, confirming our conjecture that the discriminative policy tends to overfit to environments that were seen during training.

% %Following \citet{r4r} and \citet{SDTW}, we present PL, SR, SPL, CLS, nDTW and SDTW metrics.
% % We assume CLS and SDTW are the main metrics for R4R because they both consider the task success and the agent ability to follow the instruction~\citep{babywalk}. 
% We compare our models with previous fidelity-oriented agents: RCM agents~\citep{r4r}, nDTW-based agent~\citep{SDTW} and BabyWalk agent~\citep{babywalk} without the data augmentation.
% We notice that the teacher-forced generative policy agent achieves the best result in CLS.
% Our generative policy with fidelity-oriented training achieves the close performance to BabyWalk in SDTW although our model doesn't exploit curriculum-based learning of \citet{babywalk}.
% We emphasize that our generative policy is a simple approach and its neural network is the same with \citet{speaker-follower}.
% Compared with the BabyWalk agent without curriculum-based learning, our generative fidelity-oriented agent clearly outperforms it in both CLS and SDTW.

%Discriminative Policy           & 24.10 & 0.519 & 0.619 & 0.547 & 0.401 & 39.994 & 0.224 & 0.252 & 0.115 & 0.331 \\

%\midrule
%Generative+Discriminative Policy (A+B)           & 0.680 & 0.640 & 0.510 & 0.570 & 0.533 & 0.403 \\
%Generative+Discriminative Policy (A+B+BackTrack) & 0.678 & 0.636 & 0.509 & 0.564 & 0.515 & 0.397 \\
%Discriminative Policy           &  &  &  &  &  & 39.994 & 0.224 & 0.252 & 0.115 & 0.331 \\

\section{Conclusion}

We have investigated two approaches, discriminative and generative, for 
%and the existing studies exploits the discriminative approach in
the vision-and-language navigation task, and
presented the generative language-grounded policy which
%, we experimentally confirmed,
we empirically observed to perform better than the more widely used discriminative approach.
%in experiments.
We were able to combine the generative and discriminative policies and achieve the (near) state-of-the-art results for the Room-2-Room and Room-4-Room navigation datasets, despite the simplicity of both parametrization and learning relative to the existing baselines. 
Finally, we have demonstrated that the proposed generative approach is more interpretable than discriminative ones by designing a token-wise prediction entropy.

%In future, we apply our generative language-grounded approach and 1-TENT visualization for related vision-and-language tasks.
%Finally, we design a metric for visualizing the behavior of the generative language-grounded policy and demonstrate its effectiveness.
The proposed generative parameterization, including 1-TENT visualization, is directly applicable to language-grounded reinforcement learning, such as \cite{Zhong2020RTFM:,gll}, which should be investigated in the future.
%We are also interested in the case that the
The proposed generative parametrization further enables natural integration of large-scale language model pretraining, such as \cite{gpt2,gpt3}, for various language-conditioned tasks.
% is combined with a , which should be also investigated,
%1-TENT visualization become a string tool to analyze agent behaviors for these studies. 
It is however important to investigate an efficient way to approximate the posterior distribution in order to cope with a large action set, for instance, by importance sampling and amortized inference, for the proposed generative parametrization to be more broadly applicable, in the future.

%We proposed the speaker-driven exploration approach, which avoid models to overfitting on seen environments and allows to follow instructions especially in unseen environments.
%Hopefully our analyses in this paper become a strong tool in develop models that think as humans and then act. We also hope this become a bit useful tool to realize how we use our languages in our higher abilities in decision making.

\subsubsection*{Acknowledgments}
SK was supported by ACT-I, JST.
KC was supported by NSF Award 1922658 NRT-HDR: FUTURE Foundations, Translation.

\bibliography{vln_paper}

\begin{thebibliography}{32}
\providecommand{\natexlab}[1]{#1}
\providecommand{\url}[1]{\texttt{#1}}
\expandafter\ifx\csname urlstyle\endcsname\relax
  \providecommand{\doi}[1]{doi: #1}\else
  \providecommand{\doi}{doi: \begingroup \urlstyle{rm}\Url}\fi

\bibitem[Anderson et~al.(2018{\natexlab{a}})Anderson, Chang, Chaplot,
  Dosovitskiy, Gupta, Koltun, Kosecka, Malik, Mottaghi, Savva, and Zamir]{spl}
Peter Anderson, Angel~X. Chang, Devendra~Singh Chaplot, Alexey Dosovitskiy,
  Saurabh Gupta, Vladlen Koltun, Jana Kosecka, Jitendra Malik, Roozbeh
  Mottaghi, Manolis Savva, and Amir~Roshan Zamir.
\newblock On evaluation of embodied navigation agents.
\newblock \emph{ArXiv}, abs/1807.06757, 2018{\natexlab{a}}.

\bibitem[Anderson et~al.(2018{\natexlab{b}})Anderson, Wu, Teney, Bruce,
  Johnson, S{\"u}nderhauf, Reid, Gould, and van~den Hengel]{mattersim}
Peter Anderson, Qi~Wu, Damien Teney, Jake Bruce, Mark Johnson, Niko
  S{\"u}nderhauf, Ian Reid, Stephen Gould, and Anton van~den Hengel.
\newblock {Vision-and-Language Navigation}: Interpreting visually-grounded
  navigation instructions in real environments.
\newblock In \emph{Proceedings of the IEEE Conference on Computer Vision and
  Pattern Recognition (CVPR)}, 2018{\natexlab{b}}.

\bibitem[Bahdanau et~al.(2014)Bahdanau, Cho, and Bengio]{bahdanau2014neural}
Dzmitry Bahdanau, Kyunghyun Cho, and Yoshua Bengio.
\newblock Neural machine translation by jointly learning to align and
  translate, 2014.
\newblock URL \url{http://arxiv.org/abs/1409.0473}.

\bibitem[Brown et~al.(2020)Brown, Mann, Ryder, Subbiah, Kaplan, Dhariwal,
  Neelakantan, Shyam, Sastry, Askell, Agarwal, Herbert-Voss, Krueger, Henighan,
  Child, Ramesh, Ziegler, Wu, Winter, Hesse, Chen, Sigler, Litwin, Gray, Chess,
  Clark, Berner, McCandlish, Radford, Sutskever, and Amodei]{gpt3}
Tom~B. Brown, Benjamin Mann, Nick Ryder, Melanie Subbiah, Jared Kaplan,
  Prafulla Dhariwal, Arvind Neelakantan, Pranav Shyam, Girish Sastry, Amanda
  Askell, Sandhini Agarwal, Ariel Herbert-Voss, Gretchen Krueger, Tom Henighan,
  Rewon Child, Aditya Ramesh, Daniel~M. Ziegler, Jeffrey Wu, Clemens Winter,
  Christopher Hesse, Mark Chen, Eric Sigler, Mateusz Litwin, Scott Gray,
  Benjamin Chess, Jack Clark, Christopher Berner, Sam McCandlish, Alec Radford,
  Ilya Sutskever, and Dario Amodei.
\newblock Language models are few-shot learners.
\newblock 2020.

\bibitem[Chang et~al.(2017)Chang, Dai, Funkhouser, Halber, Niessner, Savva,
  Song, Zeng, and Zhang]{Matterport3D}
Angel Chang, Angela Dai, Thomas Funkhouser, Maciej Halber, Matthias Niessner,
  Manolis Savva, Shuran Song, Andy Zeng, and Yinda Zhang.
\newblock {Matterport3D}: Learning from {RGB-D} data in indoor environments.
\newblock \emph{International Conference on 3D Vision (3DV)}, 2017.

\bibitem[Chen et~al.(2019)Chen, Suhr, Misra, Snavely, and Artzi]{touchdown}
Howard Chen, Alane Suhr, Dipendra Misra, Noah Snavely, and Yoav Artzi.
\newblock Touchdown: Natural language navigation and spatial reasoning in
  visual street environments.
\newblock In \emph{Proceedings of the IEEE/CVF Conference on Computer Vision
  and Pattern Recognition (CVPR)}, June 2019.

\bibitem[Das et~al.(2018)Das, Datta, Gkioxari, Lee, Parikh, and
  Batra]{embodiedqa}
Abhishek Das, Samyak Datta, Georgia Gkioxari, Stefan Lee, Devi Parikh, and
  Dhruv Batra.
\newblock {E}mbodied {Q}uestion {A}nswering.
\newblock In \emph{Proceedings of the IEEE Conference on Computer Vision and
  Pattern Recognition (CVPR)}, 2018.

\bibitem[Fried et~al.(2018)Fried, Hu, Cirik, Rohrbach, Andreas, Morency,
  Berg-Kirkpatrick, Saenko, Klein, and Darrell]{speaker-follower}
Daniel Fried, Ronghang Hu, Volkan Cirik, Anna Rohrbach, Jacob Andreas,
  Louis-Philippe Morency, Taylor Berg-Kirkpatrick, Kate Saenko, Dan Klein, and
  Trevor Darrell.
\newblock Speaker-follower models for vision-and-language navigation.
\newblock In S.~Bengio, H.~Wallach, H.~Larochelle, K.~Grauman, N.~Cesa-Bianchi,
  and R.~Garnett (eds.), \emph{Advances in Neural Information Processing
  Systems 31}, pp.\  3314--3325. Curran Associates, Inc., 2018.
\newblock URL
  \url{http://papers.nips.cc/paper/7592-speaker-follower-models-for-vision-and-language-navigation.pdf}.

\bibitem[Gordon et~al.(2018)Gordon, Kembhavi, Rastegari, Redmon, Fox, and
  Farhadi]{iqa}
Daniel Gordon, Aniruddha Kembhavi, Mohammad Rastegari, Joseph Redmon, Dieter
  Fox, and Ali Farhadi.
\newblock {IQA}: Visual question answering in interactive environments.
\newblock In \emph{2018 {IEEE}/{CVF} Conference on Computer Vision and Pattern
  Recognition}. {IEEE}, June 2018.
\newblock \doi{10.1109/cvpr.2018.00430}.
\newblock URL \url{https://doi.org/10.1109/cvpr.2018.00430}.

\bibitem[Hao et~al.(2020)Hao, Li, Li, Carin, and Gao]{prevalent}
Weituo Hao, Chunyuan Li, Xiujun Li, Lawrence Carin, and Jianfeng Gao.
\newblock Towards learning a generic agent for vision-and-language navigation
  via pre-training.
\newblock volume abs/2002.10638, 2020.
\newblock URL \url{https://arxiv.org/abs/2002.10638}.

\bibitem[He et~al.(2016)He, Zhang, Ren, and Sun]{resnet}
Kaiming He, Xiangyu Zhang, Shaoqing Ren, and Jian Sun.
\newblock Deep residual learning for image recognition.
\newblock In \emph{2016 {IEEE} Conference on Computer Vision and Pattern
  Recognition, {CVPR} 2016, Las Vegas, NV, USA, June 27-30, 2016}, pp.\
  770--778. {IEEE} Computer Society, 2016.
\newblock \doi{10.1109/CVPR.2016.90}.
\newblock URL \url{https://doi.org/10.1109/CVPR.2016.90}.

\bibitem[Hermann et~al.(2017)Hermann, Hill, Green, Wang, Faulkner, Soyer,
  Szepesvari, Czarnecki, Jaderberg, Teplyashin, Wainwright, Apps, Hassabis, and
  Blunsom]{gll}
Karl~Moritz Hermann, Felix Hill, Simon Green, Fumin Wang, Ryan Faulkner, Hubert
  Soyer, David Szepesvari, Wojciech~Marian Czarnecki, Max Jaderberg, Denis
  Teplyashin, Marcus Wainwright, Chris Apps, Demis Hassabis, and Phil Blunsom.
\newblock Grounded language learning in a simulated 3d world.
\newblock \emph{CoRR}, abs/1706.06551, 2017.
\newblock URL \url{http://arxiv.org/abs/1706.06551}.

\bibitem[Hu et~al.(2019)Hu, Fried, Rohrbach, Klein, Darrell, and
  Saenko]{vln-areyoulooking}
Ronghang Hu, Daniel Fried, Anna Rohrbach, Dan Klein, Trevor Darrell, and Kate
  Saenko.
\newblock Are you looking? grounding to multiple modalities in
  vision-and-language navigation.
\newblock In \emph{Proceedings of the 57th Annual Meeting of the Association
  for Computational Linguistics}, pp.\  6551--6557, Florence, Italy, July 2019.
  Association for Computational Linguistics.
\newblock \doi{10.18653/v1/P19-1655}.
\newblock URL \url{https://www.aclweb.org/anthology/P19-1655}.

\bibitem[Huang et~al.(2019)Huang, Jain, Mehta, Ku, Magalh{\~{a}}es, Baldridge,
  and Ie]{trl-vln}
Haoshuo Huang, Vihan Jain, Harsh Mehta, Alexander Ku, Gabriel Magalh{\~{a}}es,
  Jason Baldridge, and Eugene Ie.
\newblock Transferable representation learning in vision-and-language
  navigation.
\newblock In \emph{2019 {IEEE/CVF} International Conference on Computer Vision,
  {ICCV} 2019, Seoul, Korea (South), October 27 - November 2, 2019}, pp.\
  7403--7412. {IEEE}, 2019.
\newblock \doi{10.1109/ICCV.2019.00750}.
\newblock URL \url{https://doi.org/10.1109/ICCV.2019.00750}.

\bibitem[Ilharco et~al.(2019)Ilharco, Jain, Ku, Ie, and Baldridge]{SDTW}
Gabriel Ilharco, Vihan Jain, Alexander Ku, Eugene Ie, and Jason Baldridge.
\newblock General evaluation for instruction conditioned navigation using
  dynamic time warping.
\newblock In \emph{Visually Grounded Interaction and Language (ViGIL), NeurIPS
  2019 Workshop, Vancouver, Canada, December 13, 2019}, 2019.
\newblock URL \url{https://vigilworkshop.github.io/static/papers/33.pdf}.

\bibitem[Jain et~al.(2019)Jain, Magalhaes, Ku, Vaswani, Ie, and Baldridge]{r4r}
Vihan Jain, Gabriel Magalhaes, Alexander Ku, Ashish Vaswani, Eugene Ie, and
  Jason Baldridge.
\newblock Stay on the path: Instruction fidelity in vision-and-language
  navigation.
\newblock In \emph{Proceedings of the 57th Annual Meeting of the Association
  for Computational Linguistics}, pp.\  1862--1872, Florence, Italy, July 2019.
  Association for Computational Linguistics.
\newblock \doi{10.18653/v1/P19-1181}.
\newblock URL \url{https://www.aclweb.org/anthology/P19-1181}.

\bibitem[Ke et~al.(2019)Ke, Li, Bisk, Holtzman, Gan, Liu, Gao, Choi, and
  Srinivasa]{cvpr2019_fast}
Liyiming Ke, Xiujun Li, Yonatan Bisk, Ari Holtzman, Zhe Gan, Jingjing Liu,
  Jianfeng Gao, Yejin Choi, and Siddhartha Srinivasa.
\newblock Tactical rewind: Self-correction via backtracking in
  vision-and-language navigation.
\newblock In \emph{The IEEE Conference on Computer Vision and Pattern
  Recognition (CVPR)}, June 2019.
\newblock URL
  \url{http://openaccess.thecvf.com/content_CVPR_2019/html/Ke_Tactical_Rewind_Self-Correction_via_Backtracking_in_Vision-And-Language_Navigation_CVPR_2019_paper.html}.

\bibitem[Li et~al.(2019)Li, Li, Xia, Bisk, Celikyilmaz, Gao, Smith, and
  Choi]{vln-press}
Xiujun Li, Chunyuan Li, Qiaolin Xia, Yonatan Bisk, Asli Celikyilmaz, Jianfeng
  Gao, Noah~A. Smith, and Yejin Choi.
\newblock Robust navigation with language pretraining and stochastic sampling.
\newblock In \emph{Proceedings of the 2019 Conference on Empirical Methods in
  Natural Language Processing and the 9th International Joint Conference on
  Natural Language Processing (EMNLP-IJCNLP)}, pp.\  1494--1499, Hong Kong,
  China, November 2019. Association for Computational Linguistics.
\newblock \doi{10.18653/v1/D19-1159}.
\newblock URL \url{https://www.aclweb.org/anthology/D19-1159}.

\bibitem[Ma et~al.(2019)Ma, Lu, Wu, AlRegib, Kira, Socher, and Xiong]{vln-sm}
Chih{-}Yao Ma, Jiasen Lu, Zuxuan Wu, Ghassan AlRegib, Zsolt Kira, Richard
  Socher, and Caiming Xiong.
\newblock Self-monitoring navigation agent via auxiliary progress estimation.
\newblock In \emph{7th International Conference on Learning Representations,
  {ICLR} 2019, New Orleans, LA, USA, May 6-9, 2019}. OpenReview.net, 2019.
\newblock URL \url{https://openreview.net/forum?id=r1GAsjC5Fm}.

\bibitem[Nguyen \& Daum{\'e}~III(2019)Nguyen and Daum{\'e}~III]{hanna}
Khanh Nguyen and Hal Daum{\'e}~III.
\newblock Help, anna! visual navigation with natural multimodal assistance via
  retrospective curiosity-encouraging imitation learning.
\newblock In \emph{Proceedings of the 2019 Conference on Empirical Methods in
  Natural Language Processing and the 9th International Joint Conference on
  Natural Language Processing (EMNLP-IJCNLP)}, pp.\  684--695, Hong Kong,
  China, November 2019. Association for Computational Linguistics.
\newblock \doi{10.18653/v1/D19-1063}.
\newblock URL \url{https://www.aclweb.org/anthology/D19-1063}.

\bibitem[Nguyen et~al.(2019)Nguyen, Dey, Brockett, and Dolan]{vnla}
Khanh Nguyen, Debadeepta Dey, Chris Brockett, and Bill Dolan.
\newblock Vision-based navigation with language-based assistance via imitation
  learning with indirect intervention.
\newblock In \emph{2019 {IEEE}/{CVF} Conference on Computer Vision and Pattern
  Recognition ({CVPR})}. {IEEE}, June 2019.
\newblock \doi{10.1109/cvpr.2019.01281}.
\newblock URL \url{https://doi.org/10.1109/cvpr.2019.01281}.

\bibitem[Radford et~al.(2018)Radford, Wu, Child, Luan, Amodei, and
  Sutskever]{gpt2}
Alec Radford, Jeffrey Wu, Rewon Child, David Luan, Dario Amodei, and Ilya
  Sutskever.
\newblock Language models are unsupervised multitask learners.
\newblock 2018.
\newblock URL
  \url{https://d4mucfpksywv.cloudfront.net/better-language-models/language-models.pdf}.

\bibitem[Shridhar et~al.(2020)Shridhar, Thomason, Gordon, Bisk, Han, Mottaghi,
  Zettlemoyer, and Fox]{alfred}
Mohit Shridhar, Jesse Thomason, Daniel Gordon, Yonatan Bisk, Winson Han,
  Roozbeh Mottaghi, Luke Zettlemoyer, and Dieter Fox.
\newblock {ALFRED: A Benchmark for Interpreting Grounded Instructions for
  Everyday Tasks}.
\newblock In \emph{The IEEE Conference on Computer Vision and Pattern
  Recognition (CVPR)}, 2020.
\newblock URL \url{https://arxiv.org/abs/1912.01734}.

\bibitem[Tan et~al.(2019)Tan, Yu, and Bansal]{vln-envdrop}
Hao Tan, Licheng Yu, and Mohit Bansal.
\newblock Learning to navigate unseen environments: Back translation with
  environmental dropout.
\newblock In \emph{Proceedings of the 2019 Conference of the North {A}merican
  Chapter of the Association for Computational Linguistics: Human Language
  Technologies, Volume 1 (Long and Short Papers)}, pp.\  2610--2621,
  Minneapolis, Minnesota, June 2019. Association for Computational Linguistics.
\newblock \doi{10.18653/v1/N19-1268}.
\newblock URL \url{https://www.aclweb.org/anthology/N19-1268}.

\bibitem[Thomason et~al.(2019)Thomason, Murray, Cakmak, and Zettlemoyer]{cvdn}
Jesse Thomason, Michael Murray, Maya Cakmak, and Luke Zettlemoyer.
\newblock Vision-and-dialog navigation.
\newblock In \emph{Conference on Robot Learning 2019 (CoRL2019)}, November
  2019.
\newblock URL \url{https://arxiv.org/abs/1907.04957v3}.

\bibitem[Vaswani et~al.(2017)Vaswani, Shazeer, Parmar, Uszkoreit, Jones, Gomez,
  Kaiser, and Polosukhin]{transformer}
Ashish Vaswani, Noam Shazeer, Niki Parmar, Jakob Uszkoreit, Llion Jones,
  Aidan~N Gomez, \L~ukasz Kaiser, and Illia Polosukhin.
\newblock Attention is all you need.
\newblock In I.~Guyon, U.~V. Luxburg, S.~Bengio, H.~Wallach, R.~Fergus,
  S.~Vishwanathan, and R.~Garnett (eds.), \emph{Advances in Neural Information
  Processing Systems 30}, pp.\  5998--6008. Curran Associates, Inc., 2017.
\newblock URL
  \url{http://papers.nips.cc/paper/7181-attention-is-all-you-need.pdf}.

\bibitem[Wang et~al.(2018)Wang, Xiong, Wang, and Wang]{vln-rpa}
Xin Wang, Wenhan Xiong, Hongmin Wang, and William~Yang Wang.
\newblock Look before you leap: Bridging model-free and model-based
  reinforcement learning for planned-ahead vision-and-language navigation.
\newblock In \emph{The European Conference on Computer Vision (ECCV)},
  September 2018.
\newblock URL
  \url{https://eccv18-vlease.github.io/static/papers/look-before-you-leap.pdf}.

\bibitem[Wang et~al.(2019)Wang, Huang, {\c{C}}elikyilmaz, Gao, Shen, Wang,
  Wang, and Zhang]{vln-rcm}
Xin Wang, Qiuyuan Huang, Asli {\c{C}}elikyilmaz, Jianfeng Gao, Dinghan Shen,
  Yuan{-}Fang Wang, William~Yang Wang, and Lei Zhang.
\newblock Reinforced cross-modal matching and self-supervised imitation
  learning for vision-language navigation.
\newblock In \emph{{IEEE} Conference on Computer Vision and Pattern
  Recognition, {CVPR} 2019, Long Beach, CA, USA, June 16-20, 2019}, pp.\
  6629--6638. Computer Vision Foundation / {IEEE}, 2019.
\newblock \doi{10.1109/CVPR.2019.00679}.
\newblock URL
  \url{http://openaccess.thecvf.com/content\_CVPR\_2019/html/Wang\_Reinforced\_Cross-Modal\_Matching\_and\_Self-Supervised\_Imitation\_Learning\_for\_Vision-Language\_Navigation\_CVPR\_2019\_paper.html}.

\bibitem[Wijmans et~al.(2019)Wijmans, Datta, Maksymets, Das, Gkioxari, Lee,
  Essa, Parikh, and Batra]{eqa_matterport}
Erik Wijmans, Samyak Datta, Oleksandr Maksymets, Abhishek Das, Georgia
  Gkioxari, Stefan Lee, Irfan Essa, Devi Parikh, and Dhruv Batra.
\newblock {E}mbodied {Q}uestion {A}nswering in {P}hotorealistic {E}nvironments
  with {P}oint {C}loud {P}erception.
\newblock In \emph{Proceedings of the IEEE Conference on Computer Vision and
  Pattern Recognition (CVPR)}, 2019.

\bibitem[Wu et~al.(2018)Wu, Wu, Gkioxari, and Tian]{roomnav}
Yi~Wu, Yuxin Wu, Georgia Gkioxari, and Yuandong Tian.
\newblock Building generalizable agents with a realistic and rich 3d
  environment.
\newblock \emph{arXiv preprint arXiv:1801.02209}, 2018.

\bibitem[Zhong et~al.(2020)Zhong, Rocktäschel, and
  Grefenstette]{Zhong2020RTFM:}
Victor Zhong, Tim Rocktäschel, and Edward Grefenstette.
\newblock Rtfm: Generalising to new environment dynamics via reading.
\newblock In \emph{International Conference on Learning Representations}, 2020.
\newblock URL \url{https://openreview.net/forum?id=SJgob6NKvH}.

\bibitem[Zhu et~al.(2020)Zhu, Hu, Chen, Deng, Jain, Ie, and Sha]{babywalk}
Wang Zhu, Hexiang Hu, Jiacheng Chen, Zhiwei Deng, Vihan Jain, Eugene Ie, and
  Fei Sha.
\newblock Babywalk: Going farther in vision-and-language navigation by taking
  baby steps.
\newblock \emph{CoRR}, abs/2005.04625, 2020.
\newblock URL \url{https://arxiv.org/abs/2005.04625}.

\end{thebibliography}
\bibliographystyle{iclr2021_conference}
\clearpage
\appendix
%You may include other additional sections here.
%\section{Appendix}

\section{Background: Vision-and-Language Navigation}

In the R2R dataset~\citep{mattersim}, an agent moves on a graph that was constructed from one of the realistic 3D models of houses and buildings based on Matteport 3D dataset~\citep{Matterport3D}.
At the beginning of each trial, the agent is given textual instruction, is placed at the start node and attempts to reach at the goal node by moving along the edges.
At each node of the graph the agent observes the visual features of the surrounding environment and makes a decision to which neighbour node it will move next.
When the agent determines that the current node is sufficiently close to the destination node, it outputs ``STOP'', and the navigation trial ends.
The agent is evaluated in terms of the accuracy of their final location and the trajectory length \citep{mattersim,spl}.

The difficulties in VLN mainly arise from the diversity of textual instructions.
R2R provides multiple instructions for each trajectory. These instructions are created via crowd-sourcing, and their granularity and specificity highly vary \citep{vln-press}.
The agent furthermore needs to generalize to unseen environments.
Previous studies have reported that models with rich visual and textual features often overfit to the seen environments~\citep{vln-areyoulooking}.

\section{Relationships of Notations}
\label{sec:notations}
\begin{figure}[t]
	\hspace{0em}
    \begin{center}
	\includegraphics[scale=0.8,clip]{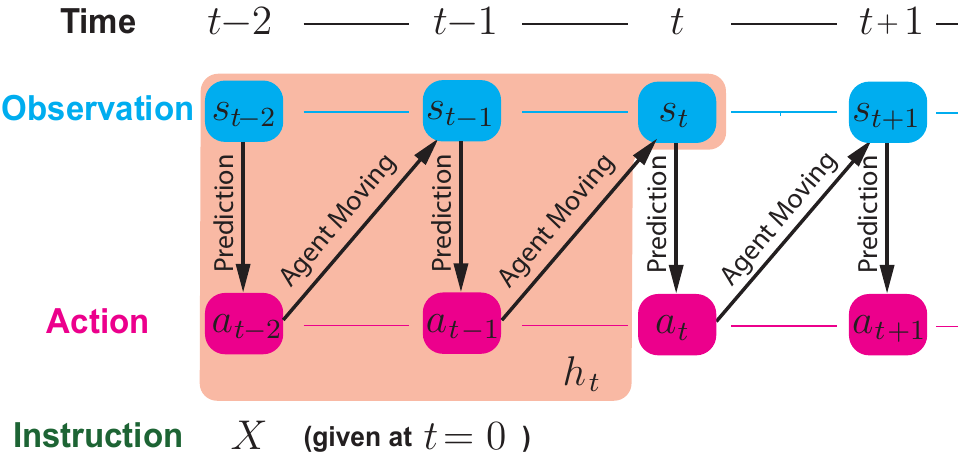}
	\vspace{-0.5em}
      \caption{
      %A graphical illustrate of a VLN policy.
			 The relation for notations of the instruction $X$, visual scene $s_t$, action $a_t$ and $h_t$ for the VLN agent.
			 }
      \label{fig:modelparams}
    \vspace{0em}
    \end{center}
\end{figure}

.
In the formalism of the generative language-grounded policy, we denote
the instruction as $X$, past and current observations as $s_{:t}$ and past actions as $a_{:t-1}$ at time step $t$.
Figure~\ref{fig:modelparams} illustrates the relationship between these notations.
$h_t$ includes the current and past observations and past actions.

\section{Details of R2R and R4R datasets}
\label{app:dataset}
R2R has in total 21,567 instructions which are 29 words long on average.
%They are created through crowd-sourcing. 
The training set has 14,025 instructions, while the validation-seen and validation-unseen datasets have 1,020 and 2,349 instructions respectively.
Each trajectory in the R2R dataset has three or four instructions.
We use the released augmentation dataset in R2R. This augmentation dataset includes 178.3K trajectories with a single instruction for each.
In the R4R dataset, training set, validation seen and validation unseen datasets contains 233k, 1k and 45k instructions respectively.\footnote{
We follow \url{https://github.com/google-research/google-research/tree/master/r4r} and generate the R4R dataset from R2R.}
We don't use augmented datasets during the R4R training.

\section{Training of Language-grounded Policies}
\label{sec:sampling}

\subsection{Fidelity-based Training for R4R Datasets}
\label{sec:r4r_fidelity}

To let the agent follow the instruction, even if the agent is out of the reference path during student-forcing learning~\citep{mattersim}, we introduce a simple heuristics to determine the reference actions.
Given the reference path $R=(r_1,...,r_i,...,r_{|R|})$ as the sequence of reference places $r_i$ and the current agent trajectory $P_t=(p_1,...,p_{t'},...,p_t)$ as the sequence of visited places $p_{t'}$ at time step $t$,
(1) if the current place $p_t$ satisfies $p_t \in R$, the reference action here is the action to follow the reference path including the stop action.\footnote{
We also consider the case that $R$ has multiple $p_t$ in the R4R training set. When an agent visits the place $p_t$ for the $n$-th time and $R$ include $m$ times of $p_t$ visiting, we assume this is the $m'$-th visiting of $x_t$ on $R$ and $m' = n$ if $n<m$ otherwise $m'=m$. We assume that the agent is at $m'$-th step of the reference path ($p_t=r_{m'}$).
}
(2) if the agent is out of the reference path at time step $t$ and was on the reference path at $t'$, we choose the temporal goal place from the reference path as $\argmin_r PL(x,R')$ where $R'=(r_i,r_{i+1},...,r_{i+t-t'})$. $PL(x,y)$ is the shortest path length between the places $x$ and $y$. Here $r_i$ is the place the agent was last on $R$ at $t'$-th step. $r_i$ is also inferred as the same way with the footnote if $R$ has multiple $x_t'$. The reference action here is the action to lead the agent to the temporal goal place in the shortest path length.

The key idea of this heuristic is that when the agent is out of the reference path, we choose the temporal goal on place from the reference path.
However, we disallow the agent to choose the temporal goal place which is far from the instruction and the visiting orders of the reference path.

%\subsection{Combination of Generative and Discriminative Policies}

\subsection{Training Details for both R2R and R4R Datasets}
%snapshot that achieves the highest SR and SPL scores in .

We use the same neural network architecture with \citet{speaker-follower}. We use minibatch-size of 25. We use a single Nvidia V100 GPU to training. We use the validation-unseen dataset to select hyperparameters.\footnote{For more details of training and evaluations, we closely follow the publicly available code \url{https://github.com/ronghanghu/speaker_follower} of \citet{speaker-follower}}

We use the mixture of supervised learning and 
%The language-grounded policy learning works with both the
imitation learning~\citep{vln-envdrop,vln-press} for both the generative and discriminative policies, which are referred as \textit{teacher-forcing} and \textit{student-forcing} \citep{mattersim}.
% We used the mixture of them for learning \cite{vln-envdrop,vln-press}.
%Especially
In particular, during training
between the reference action $a^\mathrm{T}$ and a sampled action $a^\mathrm{S}$,
we select the next action by
\begin{equation}
a=\delta a^\mathrm{S} + (1-\delta) a^\mathrm{T}
\end{equation}
where $\delta \sim \mathrm{Bernoulli}(\eta)$ following \cite{vln-press}.
We examine $\eta \in [0, 1/5, 1/3, 1/2, 1]$ using the validation set and choose $\eta=1/3$.
%We also notice that at the beginning of the navigation training, small $\eta$ contributes to the learning to follow the given path faithfully, while larger $\eta$ contributes for an agent to learn the behavior when it strays away from the path. We therefore employ linearly increasing $\eta$ during the training with the augmented dataset while we use constant $\eta$ during fintuning with only the training dataset.

After both generative and discriminative policies are trained separately, they are combined by
%The models of generative and discriminative policy are
%and a combination of them. %\textit{Explainer} and follower as of \citep{speaker-follower}.
\begin{align*}
\arg\max_{a_t} \Big\{ &\beta \log p(X|a_t,h_t) + (1-\beta) \log p_f(a_t|X,h_{t}) \Big\},
\end{align*}
to jointly navigate in the greedy experimental setting in the R2R dataset.
Here $\beta \in [0,1]$ is a hyperparameter, although
%We carefully note that
our generative model is able to navigate on itself unlike the speaker model by \cite{speaker-follower}.
$\beta$ is determined after the training of both generative and discriminative policies with the same manner.
In our experiment, we report the score of $\beta=0.5$.
%, and performs better than the discriminative policy.
%models as is experimentally shown in Sec~\ref{sec:gen_disc} and Sec.~\ref{sec:analysis_comp}.
%We discuss this in Sec~\ref{sec:gen_disc}.
% \subsection{Stochastic Sampling}
% \label{sec:sampling}

%\subsection{FAST}

%\subsection{Applying FAST for Explainer}
%\subsection{Applying FAST}
FAST \citep{cvpr2019_fast} is a framework of back-tracking to visited nodes.
%While FAST is applicable with beam decoding,
%they also propose
For single-agent back-tracking,
FAST adapts a simple heuristic to continue navigation from one of the previously visited nodes. This back-tracking is triggered when the agent visits a node second time.
Simple heuristic scoring of the sum of the transition logits is used to choose the returning node.
% is based on transition logits is
%In greedy decoding as of our experiments, FAST requires some triggers to back-tracking.
%Following this single agent back-tracking of FAST,
We use this mechanism of back-tracking in the validation and test phase.
%: when the agent revisit places where it has already visited in the final result in addition to the combination of the local speaker and follower.
%Since we need to merge model logits of the generative and discriminative policy that come from completely different loss functions, we're required to normalize their logits.
%Agent chooses the raturning place
%When the logit is negative, FAST doesn't perform well because of the fail of scoring the visited places with transition logits.
% a simple heuristic of taking
% to get a positive scores 
We use the negative inverse of the logits to determine the node to continue from each time back-tracking is triggered.
%Back-tracking is applied only in the navigation phase.
%When an agent moves back to another place due to back-tracking, this agent can pass the edges it has already traversed in the navigation graph.
All movements in back-tracking are counted in the agent trajectory and penalized in the evaluation.
%\subsection{FAST Heuristics}
%We notice that simple heuristics from FAST of \citep{cvpr2019_fast} boost the final results. FAST mainly works for beam decoding, however they also provide for single models to search their visited places as the candidates of the goal points. This searching is triggered when the single agent revisited
%FAST requires a score function based on logits to rerank the visited places. While it is straightforward to compare logits obtained from the same model,
%We don't exploit the specific rerankers that are used with beam decoding.

%\subsection{Experimental settings}
%We employ the same speaker model used in \citep{speaker-follower}.
%For follower, we use BERT-base \cite{devlin2018bert} for the encoder of the instruction text. We use the same decoder with the original Speaker-Follower model.

\section{Details of evaluation metrics}
\label{sec:metrics}
We use the following four metrics that are commonly used in evaluation for R2R navigation:
\begin{description}
\item[\textbf{Trajectory Length (TL)}] is the length of the agent trajectory in meters.
\item[\textbf{Navigation Error (NE)}] is the shortest path distance in meters from the point the agent stops to the goal point.
\item[\textbf{Success Rate (SR)}] is the proportion of successes among all the trials. The task is successful when the agent stops within 3m from the goal point \citep{mattersim}.
\item[\textbf{SPL}] is short for Success weighted by (normalized inverse) Path Length introduced in \cite{spl}. SPL is a variation of SR and is penalized by the trajectory length.
\end{description}

%We also consider the following two evaluation of nDTW and SDTW \citep{SDTW} for further analyses:
We analyze how well the trajectories followed by the proposed approach agree with the instructions using CLS~\citep{r4r}, nDTW and SDTW \citep{SDTW}. These three metrics are defined as:
\begin{description}
\item[\textbf{CLS}]  Coverage weighted by Length Score is the product of the path coverage of the reference path and the length score which penalize the longer or shorter trajectory length than the reference path length.
\item[\textbf{nDTW}] Normalized Dynamic Time Warping computes the fidelity of the trajectory given the reference path.
\item[\textbf{SDTW}] Success weighted by normalized Dynamic Time Warping is equal to nDTW for task success cases and otherwise 0.
\end{description}
In the R2R dataset, each instruction is based on the shortest path~\citep{mattersim}. The trajectory paths are specified only in the instructions and therefore these metrics evaluate how closely the models follow the instructions.  Suppose that there are two completely different routes in the navigation: the shortest path with the instruction and a different path that result in a slightly longer path length. When an agent ignores the instruction and reaches the goal on a different route, SPL will be close to 1 because of the similar path length. However, CLS and SDTW are penalized due to the completely different trajectory. 

We don't use SPL for the R4R evaluation metric because SPL depends on shortest path length of the start and goal nodes~\citep{spl} and the reference path of R4R is not the shortest path in general. In R4R, the shortest path length of the start and goal node can become 0. To consider the fidelity-based path length, we need to redefine SPL' based on the reference path length instead of the shortest path length. However, if we do so, it is incompatible to compare with SPL reported in the previous paper. Therefore we don't use SPL for R4R comparions. See \citet{r4r} for further discussions for SPL on R4R.

\begin{table*}[t]
\begin{center}
	\scriptsize\begin{tabular}{lcccccc}
        \toprule
        & \multicolumn{3}{c}{Validation (Seen)} & \multicolumn{3}{c}{Validation (Unseen)}  \\
Model                  & CLS$\uparrow$ & nDTW$\uparrow$ & SDTW$\uparrow$ & CLS$\uparrow$ & nDTW$\uparrow$ & SDTW$\uparrow$ \\
\midrule
Discriminative Policy           & 0.619 & 0.588 & 0.445 & 0.488 & 0.458 & 0.304 \\
Discriminative Policy +Aug. (A) & 0.633 & 0.596 & 0.445 & 0.528 & 0.496 & 0.356 \\
Generative Policy               & 0.625 & 0.579 & 0.427 & 0.514 & 0.478 & 0.344 \\
Generative Policy +Aug. (B)     & \textbf{0.664} & \textbf{0.630} & \textbf{0.505} & \textbf{0.539} & \textbf{0.503} & \textbf{0.379} \\
\midrule
Generative+Discriminative Policy (A+B)           & 0.680 & 0.640 & 0.510 & 0.570 & 0.533 & 0.403 \\
Generative+Discriminative Policy (A+B+BackTrack) & 0.678 & 0.636 & 0.509 & 0.564 & 0.515 & 0.397 \\
\bottomrule
	\end{tabular}
    \caption{CLS, nDTW and SDTW for \textit{generative policies} and \textit{discriminative policies} in R2R validation seen and validation unseen sets.
    Models are the same with described in Table~\ref{table:speakervsfollower}.
    Bold fonts are used for the best result as a single model.
    }
    %\vspace{-0.5em}
    \label{table:sdtw}
\end{center}
\end{table*}

\section{Details of Baseline models}
\label{sec:baselines}

\subsection{R2R Baselines}
We compare our approach against the following previous baselines in R2R.
%We briefly introduce them in order of their appearances.
%in the single run experiments.
All of these, except for the random agent, follow the discriminative approach.
% and exploit some language encoders in their neural network.
\begin{description}
\item[\textbf{Random}] An agent that moves to one random direction for five steps \citep{mattersim}.
\item[\textbf{Seq2Seq}] An LSTM-based sequence-to-sequence model \citep{mattersim}.
\item[\textbf{RPA}] Combination of model-free and model-based reinforcement learning with a look-ahead module \citep{vln-rpa}.
\item[\textbf{Follower}] An agent with panoramic view and trained with data augmentation~\citep{speaker-follower}.
%, their result comes from the follower .
\item[\textbf{Self-Monitoring}] An agent that integrates visual and textual matching trained with progress monitor regularizer \citep{vln-sm}.
\item[\textbf{RCM}] An agent that enforces cross-modal grounding of language and vision features \citep{vln-rcm}.
\item[\textbf{EnvDrop}] An agent trained with combination of imitation learning and reinforcement learning after pretraining using environmental dropout and back translation for environmental data augmentation \citep{vln-envdrop}.
\item[\textbf{FAST}] An agent that exploits the fusion score of the local action selector and the progress monitor. This agent is able to back-track to visited nodes \citep{cvpr2019_fast}.
\item[\textbf{PRESS}] An agent with the pretrained language encoder of BERT and the capability to incorporate multiple introductions for one trajectory~\citep{vln-press}. We compare our model against their model trained with a single instruction.
%introduction.
\end{description}

\subsection{R4R Baselines}
We compare our policies with models that are trained without augmented data if they are available.
\begin{description}
\item[\textbf{RCM}] An RCM agent that enforces cross-modal grounding of language and vision features reported in \citet{r4r}.
\item[\textbf{nDTW}] An agent that with reinforcement learning with nDTW-based rewards \citep{SDTW}.
\item[\textbf{BabyWalk}] An agent that exploits the proposed BABY-STEPs to follow micro-instructions. BABY-STEPs are shorter navigation tasks and trained with other learning regimes such as imitation learning (IL), reinforcement learning (RL) and curriculum reinforcement learning (Cur.).
%introduction.
\end{description}
R4R is a dataset to measure the agent fidelity to the given instruction. Therefore we choose the fidelity-oriented agents in comparison and we develop our policies with supervised learning or fidelity-oriented training.

% \section{Trajectory Fidelity in R2R Dataset}
% \label{sec:r2r_cls}

% Table~\ref{table:sdtw} shows CLS~\citep{r4r}, nDTW and SDTW~\citep{SDTW} for both generative and discriminative policies. The generative policy achieves higher CLS, nDTW and SDTW than the discriminative policy does, which suggests that the proposed generative policy follows the reference path more closely compared to the discriminative one. We conjecture this is because the generative policy is sensitive to the language instructions by construction.
% %The discriminative policy, however, may overlook the clue instructions.

\section{Agreement of generative and discriminative policies}
\label{app:polices_diff}
We present the agreement rate of action prediction between generative and discriminative policies at the bottom of Figure~\ref{fig:prec_action}.
The agreement drops over time, which implies that these policies behaves differently from each other, capturing different aspects of VLN.
The agreement become lower in later time steps and we hence consider the combination of Gen and Disc can work better than either model.

%It seems that the navigation becomes difficult with larger $t$ especially in unseen environments. The blue  solid lines of Gen is above the red lines of Disc in all time steps in unseen environment. However the result of Disc become more accurate with a large $t$ in seen environments.
%We assume this is because the Disc model is tend to quit navigation and output ``STOP''.

\begin{figure*}[t]
% \begin{minipage}{0.5\hsize}
% \begin{center}
% \vspace{-1em}
% \includegraphics[scale=0.45,clip]{fig/accu_Seen_pub.pdf}
% \end{center}
% \end{minipage}
% \begin{minipage}{0.5\hsize}
% \begin{center}
% \vspace{-1em}
% \hspace{-2em}
% \includegraphics[scale=0.45]{fig/accu_Unseen_pub.pdf}
% \end{center}
% \end{minipage}
%\vspace{-1em}
%%%%%%%%%
\begin{minipage}{0.5\hsize}
\begin{center}
\vspace{-1em}
\includegraphics[scale=0.45,clip]{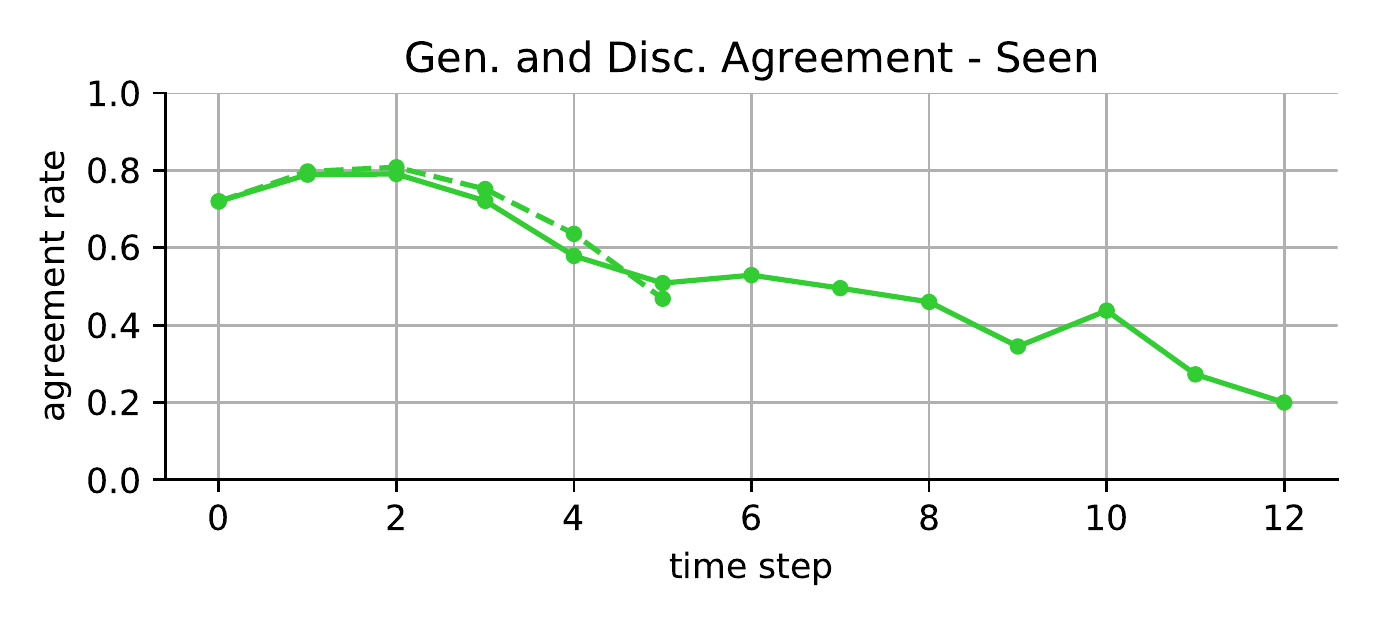}
\end{center}
\end{minipage}
\begin{minipage}{0.5\hsize}
\begin{center}
\vspace{-1em}
\hspace{-2em}
\includegraphics[scale=0.45,clip]{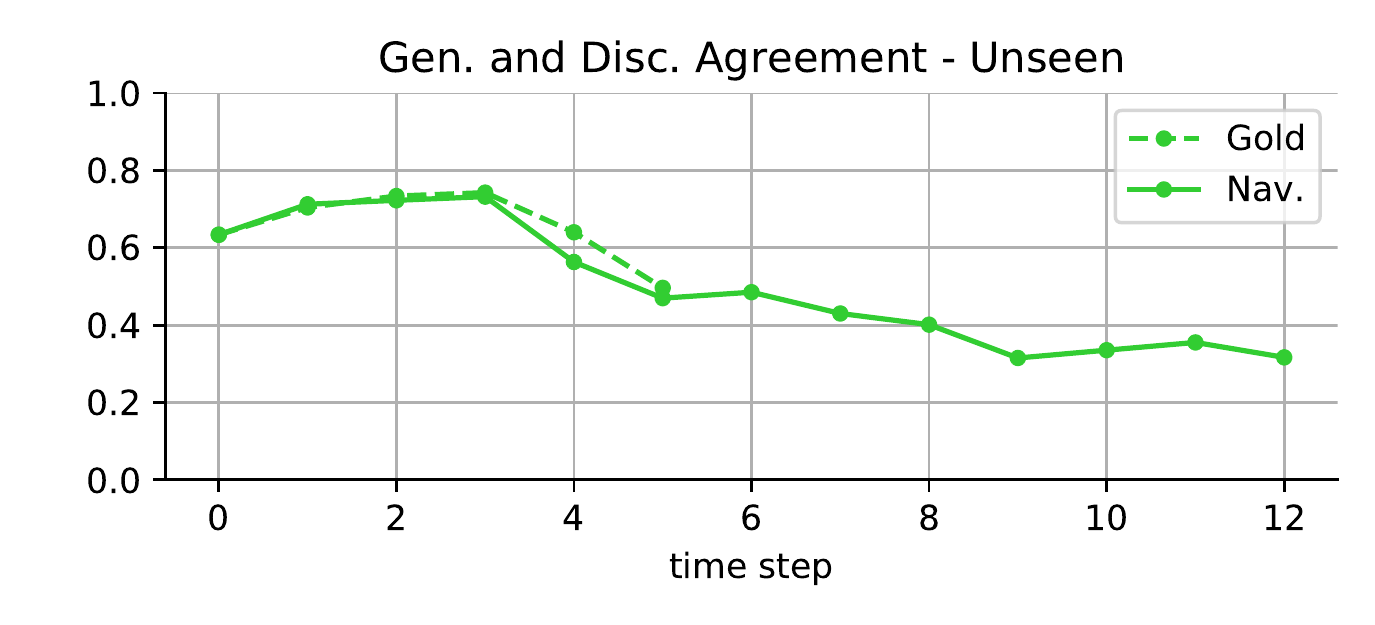}
\end{center}
\end{minipage}
\vspace{-1em}
\caption{
%The precision of actions by the generative (red) and discriminative (blue) models on the reference trajectory (dashed lines) and on navigation trajectories (solid lines).
%Solid lines are accuracy for all actions. Dotted lines are accuracy of actions when the correct actions are ``STOP'' at that place, while the dashed lines are accuracy of actions when the correct actions are not ``STOP'' at that place.
%\textbf{Bottom}:
The agreement of actions between the generative and discriminative models  on shortest paths (dashed lines) and on navigation trials (solid lines). The horizontal axis corresponds to the time step of trials.
}
\label{fig:prec_action}
\end{figure*}
% \appendix
% \section{Appendix}

\end{document}